%% file: main.tex
\documentclass[journal]{IEEEtran}
\usepackage[T1]{fontenc}
\usepackage{times}

\usepackage{amsmath}

\usepackage{url}
\usepackage{hyperref}
\hypersetup{hidelinks}
\usepackage{fvextra}
\RecustomVerbatimEnvironment{verbatim}{Verbatim}{breaklines,breakanywhere,fontsize=\tiny}
\usepackage[numbers,sort&compress]{natbib}
\usepackage{booktabs}
\usepackage{graphicx}
\usepackage{subcaption}
\usepackage{multirow}
\usepackage{array}
\usepackage{tabularx}
\usepackage[table]{xcolor}
\definecolor{sectionpurple}{RGB}{102,51,153}

\usepackage{enumitem}
\setlist{topsep=2pt,itemsep=1pt,parsep=0pt,partopsep=0pt}
\usepackage{tcolorbox}
\tcbuselibrary{breakable,skins}

\definecolor{promptpurple}{RGB}{220,210,240}
\definecolor{promptgray}{RGB}{245,245,247}

\definecolor{grpblue}{RGB}{236,241,249}   %
\definecolor{grpgreen}{RGB}{238,246,238}  %
\definecolor{grpamber}{RGB}{252,246,233}  %
\newtcolorbox{promptbox}[1]{
  enhanced,
  breakable,
  colback=promptgray,
  colframe=promptpurple,
  boxrule=0.8pt,
  arc=4pt,
  title=#1,
  fonttitle=\bfseries\small,
  coltitle=black,
  colbacktitle=promptpurple,
  colframe=promptpurple,
  titlerule=0.8pt,
  toptitle=5pt,
  bottomtitle=5pt,
  top=6pt,
  left=6pt,
  right=6pt,
  bottom=6pt,
}

\title{GameReplica: A Benchmark for Black-Box Visual Game Replication by Vision-Language Agents}

\author{Boyu Qiao, Zixin Tang, Kaige Li, Xiaoshuai Hao, Dongyu She and Xiaochun Cao%
\thanks{Boyu Qiao is with the Institute of Information Engineering, Chinese Academy of Sciences, and the School of Cyber Security, University of Chinese Academy of Sciences, Beijing, China. E-mail: \{qiaoboyu\}@iie.ac.cn.}
\thanks{Zixin Tang and Dongyu She are with Zhongguancun Laboratory, Beijing, China. E-mail: \{tangzixin, shedy\}@zgclab.edu.cn.}
\thanks{Kaige Li and Xiaochun Cao are with School of Cyberspace Security, Sun Yat-sen University.}
\thanks{Xiaoshuai Hao is with Xiaomi EV, Beijing, China. E-mail: \{haoxiaoshuai714\}@163.com.}
}

\newlength{\processfigheight}
\newsavebox{\fittablebox}
\newcommand{\fittable}[1]{%
  \sbox{\fittablebox}{#1}%
  \ifdim\wd\fittablebox>\textwidth
    \resizebox{\textwidth}{!}{\usebox{\fittablebox}}%
  \else
    \usebox{\fittablebox}%
  \fi}

\begin{document}
\maketitle

\begin{abstract}
Benchmarks for coding agents typically assess implementation from explicit specifications, leaving open whether agents can discover and reproduce systems whose rules are undocumented and accessible only through interaction. We study this capability in web games, which offer controlled environments with observable actions and outcomes, and introduce GameReplica, a benchmark for end-to-end black-box game replication. It comprises 125 instances spanning 25 web games and five core mechanism families, with each game instantiated at five difficulty levels. Under container isolation, an agent accesses the target game only through an action--screenshot interface, may iterate between probing the game, implementing a replica, and checking it, and delivers a self-contained runnable replica together with a rule checklist. A reference-grounded evaluation then scores the deliverables separately on rule consistency, implementation consistency, and visual fidelity. Experiments across six models and five agent runtimes show that black-box replication remains challenging: Claude Opus 4.8 achieves the highest Overall score, 71.6\%, while the other models obtain 4.0\%--43.0\%. Every model scores highest on visual fidelity and lowest on rule consistency, suggesting that reproducing appearance is easier than recovering mechanics. Difficulty widens the gap: from L1 to L5, the four lower-scoring models show relative Overall score declines of 33\%--63\%, whereas the strongest model declines by less than 5\%. In the evaluated configurations, outcomes vary more across models than across agent runtimes, and most attributed failures of weaker models arise during exploration, before replica code is written. Resources and interactive demonstrations are available at \url{https://qqqqqqby.github.io/GameReplica/}.
\end{abstract}

\begin{IEEEkeywords}
Coding agents, vision-language models, benchmark, specification recovery, web games, agent evaluation.
\end{IEEEkeywords}

\section{Introduction}

\IEEEPARstart{C}{oding} agents powered by Vision-Language Models (VLMs) have rapidly transitioned from simple code generators into sophisticated systems capable of repository refactoring, program execution, and end-to-end application synthesis \citep{jimenez2024swebench,wang2025uitars2,bandel2026generalagent}. As these agents are increasingly integrated into real-world software engineering workflows, rigorous benchmarks are needed to characterize the capabilities and limitations of current models.
However, existing benchmarks typically rely on target behaviors predefined via textual requirements, issue descriptions, design mockups, or existing code. Consequently, these environments evaluate specification implementation rather than autonomous specification recovery. We address this gap by investigating a paradigm where the target system is available solely as a running black box. 
Our central inquiry is whether an agent can deduce these latent mechanics through active visual probing and translate them into a functionally equivalent program.

\begin{figure*}[t]
\centering
\includegraphics[width=0.95\textwidth]{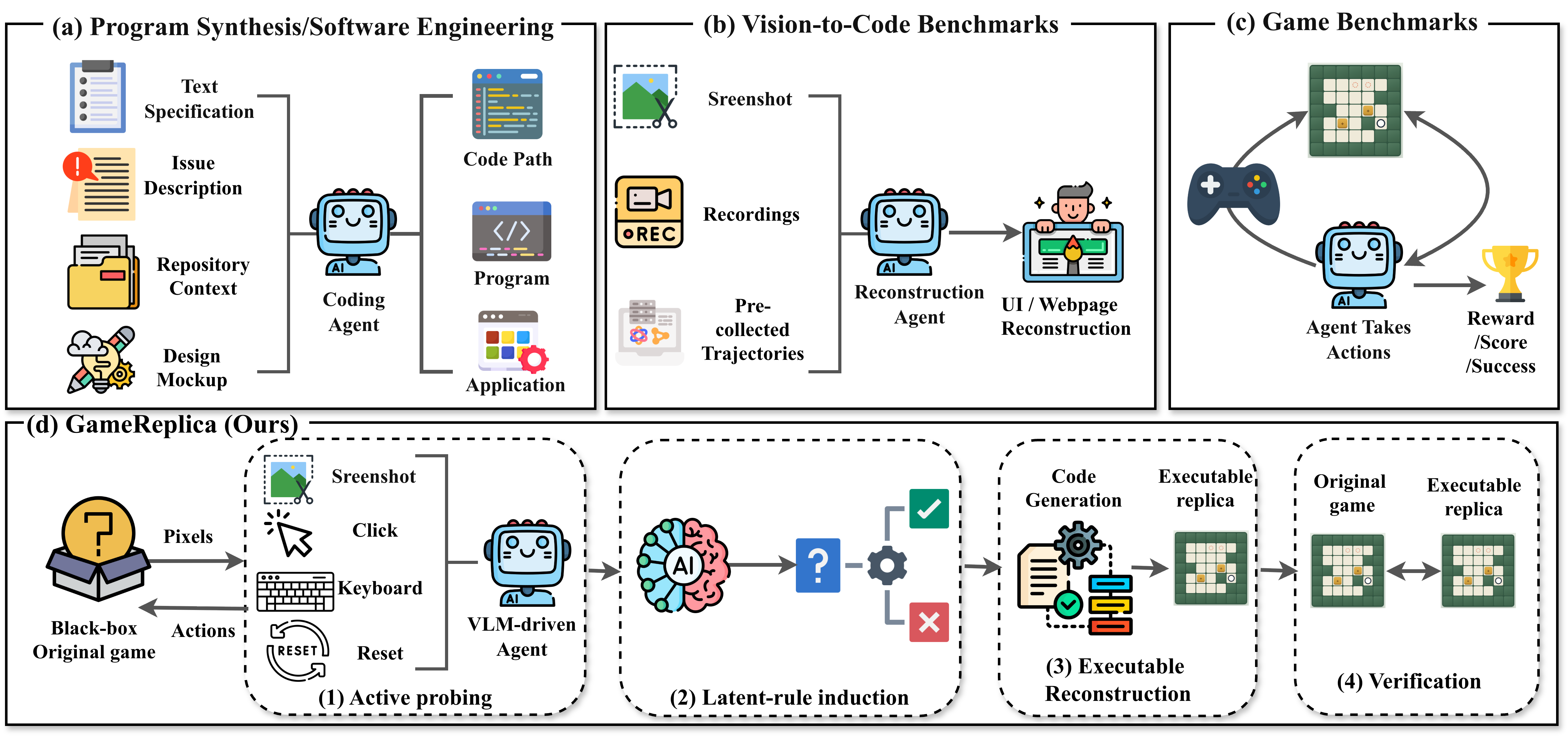}
\caption{Comparison of existing benchmarks and GameReplica.
(a) Program synthesis and software engineering benchmarks provide explicit specifications for implementation.
(b) Vision-to-code benchmarks provide visual observations or pre-collected interaction trajectories.
(c) Game benchmarks evaluate agent performance in predefined environments without requiring executable reconstruction.
(d) GameReplica instead requires agents to actively probe black-box games, infer latent rules, reconstruct executable replicas, and verify their behavior against the original games.}
\label{fig:intro_comparison}
\end{figure*}

Prior research along this line can be broadly classified into three paradigms, yet none addresses the full process of autonomous specification recovery, as illustrated in Figure~\ref{fig:intro_comparison}.
Conventional program-synthesis and software-engineering benchmarks supply explicit target behaviors via issue descriptions and repository contexts \citep{jimenez2024swebench,yang2025swebenchmm}, textual requirements and design mockups \citep{lu2025webgenbench,guo2026vista}, or predefined game projects \citep{chi2026gamedevbench,la2026gameenginebench,sun2026jamer}. 
Vision-to-code
benchmarks remove some textual supervision, yet static settings expose
appearance without interaction \citep{si2024design2code,yang2025ui2coden}, and
interactive variants still provide descriptions, recordings, or pre-collected
trajectories \citep{xiao2025interaction2code,chen2025iwrbench}. Game benchmarks
evaluate decision making and exploration within a fixed environment
\citep{paglieri2024balrog,hu2025lmgamebench,zhang2025videogamebench,jiang2026gbqa},
but do not require the inferred mechanics to be reimplemented. 
Ultimately, these prior settings pre-supply the underlying specification, the interaction trajectory, or the execution environment itself. Consequently, they do not jointly evaluate active visual probing, latent-rule induction, executable code synthesis, and external evaluation of the result.

To address this gap, we introduce \textbf{GameReplica} to evaluate autonomous specification recovery within a controlled black-box setting.
The benchmark comprises 125 instances spanning 25 web games and five core mechanism families, systematically scaled across five difficulty levels per game. Operating only through an action--screenshot interface, agents capture screenshots, execute keyboard and mouse inputs, trigger resets, and probe boundary conditions. They have no access to reference implementations, test suites, explicit specifications, internal states, or answer files.
To replicate a game faithfully, an agent must recover its visual assets, state transitions, action constraints, and termination conditions, and then synthesize them into a self-contained executable replica.
Structurally, \textbf{GameReplica} is anchored by three core modules encompassing complexity-controlled instance construction, pixel-based interaction, and reference-grounded evaluation. Each game family follows a component-wise complexity model, with semantically aligned specifications, rule checklists, and reference implementations at every level. Separate measures of rule consistency, implementation consistency, and visual fidelity distinguish induced rules, implemented mechanics, and rendered appearance.

Experiments reveal a clear gap between visual resemblance and the faithful reconstruction of game mechanics. The best-performing agent achieves an Overall score of 71.6\%, while the remaining agents obtain 4.0\%--43.0\%. Lower-performing agents score 7.2\%--62.8\% on visual fidelity and 3.6\%--39.4\% on implementation consistency, so their visual fidelity scores are consistently higher than their implementation consistency scores. The performance gap widens with difficulty as the four lower-scoring models show relative Overall score declines of 33\%--63\% from L1 to L5, compared with about 5\% for the best-performing model. Our contributions are as follows:

\begin{itemize}

  \item We formulate black-box game replication as specification recovery by coding agents. Given only pixel-level observations and an interactive game environment, an agent must actively probe the game to recover its visual elements and latent rules, and then implement them as a self-contained executable replica.
  
  \item We introduce \textbf{GameReplica}, a complexity-controlled benchmark for black-box game replication.
  It comprises 125 instances drawn from 25 web games that span five mechanism families and five difficulty levels, enforces black-box access, and scores rule consistency, implementation consistency, and visual fidelity separately.
  
  \item We characterize the capabilities and failure modes of current agents. Across six models and five agent runtimes, agents score higher on visual appearance than on executable mechanics, and lower-scoring models decline more as difficulty increases.
  Outcome-level and process-level analyses further show that most attributed failures of weaker models arise during exploration, before replica code is written.
\end{itemize}

\section{Related Work}
\label{sec:related}

\subsection{Program Synthesis and Software Engineering Benchmarks}
Program synthesis and software engineering benchmarks have progressively evolved from function-level code generation evaluated against isolated unit tests \citep{ivanova2026multilcb} to repository-scale issue resolution \citep{jimenez2024swebench} and multimodal frameworks incorporating screenshot contexts \citep{yang2025swebenchmm}. Application-level evaluations similarly demand that agents construct interfaces driven by explicit textual instructions \citep{lu2025webgenbench} or static visual mockups \citep{guo2026vista}. Parallel efforts in game development also supply target semantics by deriving task instances from structured tutorials \citep{chi2026gamedevbench}, assessing project-level execution within professional engines \citep{la2026gameenginebench,sun2026jamer}, or quantifying the disparity between automated systems and expert developers \citep{chen2026gamexpert}. 
Despite variations in scale and modality, these paradigms externalize the desired behavior through explicit issues, directives, tutorials, visual mockups, or predefined project architectures. 
Consequently, what they measure is the degree of realization relative to established specifications, rather than the process by which those specifications are derived from a running system.

\subsection{Vision-to-Code Benchmarks}
Vision-to-code benchmarks advance toward behavioral reconstruction by substituting textual requirements with rendered visual evidence. For static interfaces, frameworks map webpage screenshots to frontend implementations \citep{si2024design2code} or refine this mapping through iterative generation and visual comparison \citep{yang2025ui2coden}. Interactive variants introduce dynamic behavior via explicit textual descriptions and pre-recorded screen captures \citep{xiao2025interaction2code} or reconstruct webpages from user-interaction videos evaluated under an agent-as-judge protocol \citep{chen2025iwrbench}. Beyond web environments, related efforts target structured artifact synthesis by extracting layouts from rendered images \citep{roberts2024image2struct}, generating visual documents via {HTML/CSS} \citep{lin2026anydoc}, or regenerating charts directly from pixels \citep{zhang2026chart2code}. The main distinction between these frameworks and our approach lies in evidence acquisition. By furnishing screenshots, recordings, trajectories, or static assets in advance, existing settings bypass the need for agents to actively intervene and resolve latent state ambiguities. GameReplica instead makes active visual probing and evidence acquisition part of the replication task.

\subsection{Agents in Game Environments}
Games provide inherently interactive testbeds for planning, reasoning, and exploration. Recent evaluations assess agentic reasoning across diverse game suites \citep{paglieri2024balrog}, measure gameplay competence while mitigating data contamination \citep{hu2025lmgamebench}, and challenge vision-language models to navigate popular titles using raw visual inputs \citep{zhang2025videogamebench}. Beyond pure game completion, specialized benchmarks probe long-horizon exploration through injected bug discovery \citep{jiang2026gbqa} or evaluate narrative consistency within adventure games \citep{ahn2025flashadventure,ma2026interactive}. Across all these paradigms, the game serves as a fixed environment for evaluating decision-making, exploration, or task completion. GameReplica changes both the required output and the evaluation target: interaction serves to uncover latent rules, and success requires implementing the inferred mechanics as a standalone executable replica.

Taken together, prior work evaluates three adjacent capabilities: implementing
an explicit specification, reconstructing an interface from provided visual
evidence, and acting effectively in a fixed environment. Black-box replication
requires their composition under a stricter information boundary. The agent
must decide what to observe, induce a specification from interaction outcomes,
implement it, and submit the result to an external evaluation. GameReplica
scores the outcome of this end-to-end process; it does not measure each step
separately.

\section{GameReplica}

\begin{figure*}[t]
\centering
\includegraphics[width=\textwidth]{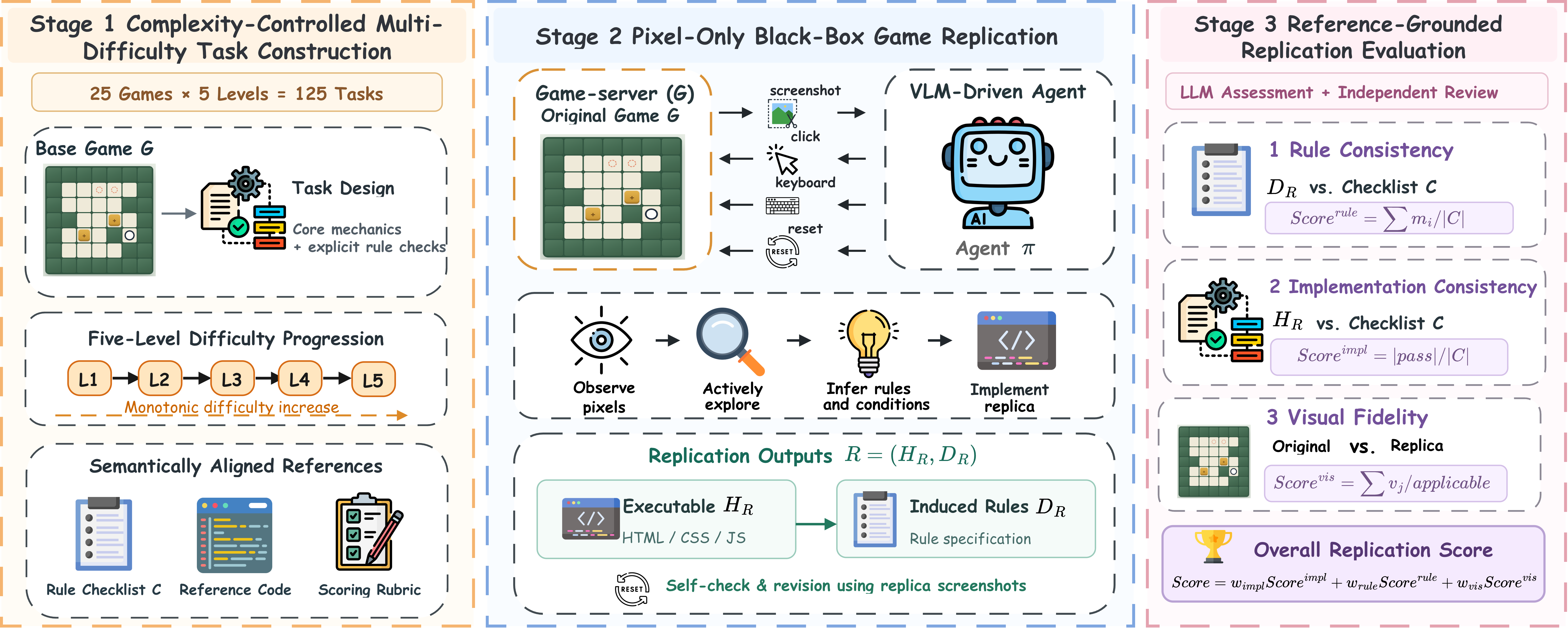}
\caption{Overview of GameReplica. \textbf{Stage 1. Complexity-Controlled Multi-Difficulty Task Construction.} Each base game is systematically scaled across five difficulty levels accompanied by a semantically aligned specification, rule checklist, and reference implementation. \textbf{Stage 2. Pixel-Only Black-Box Game Replication.} The evaluated agent actively probes the live environment only via an action--screenshot interface to synthesize a self-contained executable replica alongside an induced rule checklist. \textbf{Stage 3. Reference-Grounded Replication Evaluation.} The synthesized replica is benchmarked against curated reference artifacts across the decoupled dimensions of rule consistency, implementation consistency, and visual fidelity.}
\label{fig:overview}
\end{figure*}

GameReplica evaluates active interaction, rule induction, and executable
reconstruction through the three-stage protocol in Figure~\ref{fig:overview}.
The modules provide complexity-controlled instance construction, pixel-only
black-box replication, and reference-grounded evaluation.
First, \textit{Complexity-Controlled Multi-Difficulty Task Construction} scales each base game across five difficulty levels (L1--L5) while preserving semantic alignment among the specification, rule checklist, and reference implementation (Section~\ref{sec:gamedesign}). Second, \textit{Pixel-Only Black-Box Game Replication} restricts game access to HTTP endpoints for screenshot captures, mouse clicks, keyboard inputs, and state resets (Section~\ref{sec:gamereplication}). This constraint isolates the evaluated agent from the target source code. Finally, \textit{Reference-Grounded Replication Evaluation} compares the synthesized replica against the original game to benchmark agent capabilities across three distinct dimensions encompassing rule consistency, implementation consistency, and visual fidelity (Section~\ref{sec:replicationeval}).

\subsection{Task Definition}

This task evaluates the capacity of a VLM-driven agent to autonomously infer latent mechanics and reconstruct both the functionality and visual presentation of a web game using only visual interaction. Let $G$ denote the target game and $G^{\star}$ represent its reference implementation. The underlying rules and source code of the instance remain hidden from the agent. Consequently, interaction with $G$ occurs solely through a controlled action--screenshot interface $\mathcal{O}$ defined by the mapping
$\mathcal{O}:(a_1, a_2, \dots, a_t) \mapsto s_t$,
where $a_i$ denotes an atomic action (such as a mouse click, keystroke, or environment reset) and $s_t$ represents the screenshot rendered following the execution of the action sequence $(a_1,\dots,a_t)$. The agent has no access to $G^{\star}$, its underlying source code, and any ground-truth scoring artifacts.

The evaluated model instantiates an agent $\pi$ operating under an interaction budget $T$, enforced as a maximum execution time (Section~\ref{sec:experimental-setup}). Through the interface $\mathcal{O}$, the agent actively observes state transitions and induces latent rules to synthesize a replica $R$ formulated as
\begin{equation}
R = \pi(\mathcal{O};T), \qquad R = (H_R,\, D_R).
\end{equation}
Here, $H_R$ constitutes a self-contained single-file executable capable of running natively within a browser. Furthermore, the formalized rule checklist $D_R$ documents the agent's induced comprehension of visual assets, interaction mechanics, state transitions, and termination criteria. The task therefore provides no explicit specification: the agent must induce the rules from pixel-level feedback.

\subsection{Complexity-Controlled Multi-Difficulty Task Construction}
\label{sec:gamedesign}

This module enables controlled comparisons within each game family by
constructing five progressively harder game instances
$\{G_{L_1},\dots,G_{L_5}\}$, denoted L1--L5. Each instance contains a reference
rule checklist $C=\{c_1,\dots,c_n\}$ and source code that are kept
consistent with each other. The checklist comprises six categories of atomic
rules: visual, legal action, state, boundary, terminal lock, and level
layout. Table~\ref{tab:rule-categories} in the appendix defines these categories. We characterize a game's difficulty level using structural complexity
and solving complexity.
\begin{itemize}
\item \textbf{Structural complexity} characterizes rule complexity and state complexity.
Rule complexity counts candidate action types $N_{\text{act}}$, state-transition rules
$N_{\text{trans}}$, and termination rules $N_{\text{goal}}$.
State complexity counts state units $N_{\text{unit}}$ and dynamic object types
$N_{\text{obj}}$.
\item \textbf{Solving complexity} characterizes the minimal-solution depth
$N_{\text{step}}$ and the mean branching factor $\bar{B}$.
\end{itemize}

For any instance $G$, we define the structural complexity vector as
\begin{equation}
v_{\text{struct}}(G)=[\,N_{\text{act}},\,N_{\text{trans}},\,N_{\text{goal}},\,N_{\text{unit}},\,N_{\text{obj}}\,].
\end{equation}
The first three dimensions measure rule complexity through the numbers of
action types, state-transition rules, and goal, win/loss, and termination rules. The
last two measure state complexity through the numbers of state units and
dynamic object types. We correspondingly define the solving complexity vector
as
\begin{equation}
v_{\text{solve}}(G)=[\,N_{\text{step}},\,\bar{B}\,],
\end{equation}
where $N_{\text{step}}$ is the length of the shortest action sequence from the
initial state to the goal state, and $\bar{B}$ is the average number of legal
actions over reachable states. Solving complexity is induced by structural
complexity rather than designed independently. Enlarging the state space and
removing pre-filled information increase $N_{\text{step}}$, whereas relaxing
constraints and reducing clues weaken search pruning and increase $\bar{B}$.
Table~\ref{tab:complexity-metrics} explains the structural and solving
complexity metrics.

\textbf{Baseline and progression.}
The minimal rule set for the core gameplay defines
the lower-bound instance $G_{L_1}$. It retains only the rules required to
determine win/loss outcomes, uses the fewest state units and dynamic objects,
and provides the most complete initial pre-filling. Consequently, $G_{L_1}$ has
the shortest $N_{\text{step}}$ and the narrowest $\bar{B}$. Each subsequent
level adds complexity monotonically. During construction, each of the five
structural dimensions is assigned a direction in
$\{\text{increase},\text{fixed}\}$. Fixed dimensions remain
unchanged, while every permitted increment is mapped to a specific structural
indicator so that the source of each difficulty increase is traceable.

The progression follows a consistent pattern.
$G_{L_1}\!\to\!G_{L_2}$ expands the state space and introduces the first simple
constraint ($N_{\text{unit}}\!\uparrow,\,N_{\text{trans}}\!\uparrow$).
$G_{L_2}\!\to\!G_{L_3}$ scales the game to its standard size while leaving the
rules largely unchanged ($N_{\text{unit}}\!\uparrow$), yielding a state-dominated transition.
$G_{L_3}\!\to\!G_{L_4}$ introduces the first significant new mechanism, such as
a matching constraint, lock--key dependency, or limited-use resource
with corresponding dynamic objects and termination conditions
($N_{\text{trans}}\!\uparrow,\,N_{\text{obj}}\!\uparrow,\,N_{\text{goal}}\!\uparrow$).
This transition is rule-dominated.
$G_{L_4}\!\to\!G_{L_5}$ further enlarges the state space and adds
termination and adjudication rules, including step limits, scoring, and failure
conditions, bringing both complexity groups to their within-family peaks.

\textbf{Validity criteria.}
A five-level family
$\{G_{L_1},\dots,G_{L_5}\}$ is valid only if it satisfies three conditions
simultaneously. (i) \emph{Monotonicity}: rule complexity, state complexity, and
solving complexity each increase monotonically overall from L1 to L5.
(ii) \emph{Mechanism authenticity}: every new rule or termination condition
affects state transitions or win/loss adjudication in the reference implementation.
Documentation or visual presentation alone does not satisfy this criterion.
(iii) \emph{Rule observability}: every level's rules can be inferred through
finite visual interaction with the controlled interface.
No rule remains hidden from interaction feedback.

\subsection{Pixel-Only Black-Box Game Replication}
\label{sec:gamereplication}
\textbf{Replication under visual access constraints.}
This module enforces
visual-only access while the VLM-driven agent $\pi$ constructs a replica page
$H_R$ of the target game $G$.
The replica must match both functionality and appearance.
The agent $\pi$ accesses $G$ through an action--screenshot interface.
It receives a rendered screenshot after each click, keypress, or reset and
cannot access the source code.
Before implementing $H_R$, it must use interaction outcomes to infer $G$'s
visual presentation, gameplay rules, and termination conditions.

\textbf{Isolated execution environment.}
The black-box constraint is enforced
by the execution environment. It makes the reference implementation and
associated ground truth inaccessible through three properties.
\begin{itemize}
\item \textbf{Interface closedness.} All information the agent $\pi$ obtains
about $G$ passes through the observation interface $\mathcal{O}$.
Only the pixels exposed by $\mathcal{O}$ are accessible.
The running game instance $G$, including its source code, internal state, and
network responses, is otherwise inaccessible.
\item \textbf{Observation determinism.} The output of $\mathcal{O}$ is
determined solely by the current state of $G$ and the action sequence, and
introduces no external information through the interface.
\item \textbf{Output isolation.} The agent can read and write only its own
artifacts, and these writes cannot become a channel for reading or tampering
with $G$.
\end{itemize}
Under the above constraints, the environment contains no reachable path to the
original game's source-code ground truth. The black-box setting is therefore
enforceable and does not depend on the agent's voluntary compliance with
prompt-level instructions.

GameReplica realizes these properties through container isolation. Each
replication instance is instantiated as an independent Docker Compose project
containing two containers.
\begin{itemize}
\item \textbf{Observation service container (game-server).} This container
loads the original game $G$ in an embedded Playwright headless browser and
exposes several HTTP interfaces, such as screenshot, click, keyboard, and
reset.
\item \textbf{Agent container (agent).} This container hosts the model under
evaluation and its agent runtime, including the agent loop, tools, and working-directory
file system. The model observes the game and develops the replica code here. It can communicate with the game-server only through the HTTP
interfaces and cannot access the original game's source code, specification, or
any server-side ground truth.
\end{itemize}

Network topology provides external isolation. The bridge network
\texttt{game-net}, shared by both containers, is declared internal and has no
outbound route. The game-server is attached only to \texttt{game-net}, so its
embedded browser cannot access the public internet. The agent container is
also attached to a restricted external network (non-internal, with an outbound
allowlist) that permits traffic only to the inference gateway of the evaluated
model. Because this network provides no general internet access, the agent can
neither retrieve external knowledge nor bypass the observation interface to
access the original game.

\textbf{Replication deliverables.}
A fixed protocol requires two deliverables:
the source code of the replica page, and an atomic rule checklist summarized by
the agent. The checklist expresses the agent's induced understanding as
individually decidable propositions, allowing rule induction and implementation
to be scored independently.

\subsection{Reference-Grounded Replication Evaluation}
\label{sec:replicationeval}

This module separates inferred rules, implemented mechanics, and rendered
appearance through rule consistency, implementation consistency, and visual fidelity.
Using the notation above, let $G$ be the original game instance and
$R=(H_R, D_R)$ the agent's deliverables, where $H_R$ is the replica's source
code and $D_R$ is the induced rule checklist. Each original game
instance also comes with the reference rule checklist $C=\{c_1,\dots,c_n\}$ from
Section~\ref{sec:gamedesign}, used here as the scoring rubric, where each $c_i$
is an atomic rule. All dimensions use a cross-model assessment--review protocol.
An assessment judge assigns verdicts, and an independent review judge confirms
or overturns them.

\textbf{Implementation consistency.}
This dimension measures whether the
replica source code $H_R$ implements each rule in the reference checklist $C$.
An agent-as-judge protocol checks $H_R$ for the required mechanics.
Assessment and review agents operate in parallel across the six rule categories.
\begin{itemize}
\item \textbf{Round 1 (assessment).} An autonomous assessment agent judges
whether the replica code $H_R$ implements each rule $c_i$ in its category,
outputs a binary verdict $u_i \in \{\text{pass}, \text{fail}\}$, and cites
the corresponding code locations as evidence.
\item \textbf{Round 2 (review).} A review sub-agent receives the Round-1
verdicts and evidence, independently re-examines the source code for the same
category, and confirms or overturns each $u_i$.
Any overturned verdict is remanded for reassessment, continuing this iterative cycle until the judgment passes review or reaches a maximum cap of $K$ rounds ($K=3$ in our experiments).
\end{itemize}
The implementation consistency score is the fraction of rules that pass review:
\begin{equation}\text{Score}^{\text{impl}} = \frac{\bigl|\{\, i : u_i = \text{pass} \,\}\bigr|}{|C|}.\end{equation}

\textbf{Rule consistency.}
This dimension measures the degree of agreement
between the agent-induced rule checklist $D_R$ and the reference rule
checklist $C$. A large language model (LLM) acting as the assessment judge compares $D_R$ with $C$ item by item and
assigns each reference rule $c_i$ a verdict $m_i \in \{1, 0.5, 0\}$,
corresponding to consistent, partially consistent, and inconsistent; the review judge then confirms or overturns each verdict. The
rule consistency score is the mean of the rule-level verdicts:
\begin{equation}\text{Score}^{\text{rule}} = \frac{\sum_{i=1}^{|C|} m_i}{|C|},\end{equation}
where $|C|$ is the number of rules in the reference checklist.

\textbf{Visual fidelity.}
This dimension compares screenshots of the original
game $G$ and the replica $H_R$ using two checklists.
The game-specific visual checklist $C^{\text{vis}}$ measures whether the game
state is presented clearly and correctly.
The general art checklist $C^{\text{art}}$, shared across games, measures
stylistic properties such as color palette and typography. A multimodal LLM acting as the assessment judge receives the
screenshot pairs and both checklists and assigns each item
$e_j \in C^{\text{vis}} \cup C^{\text{art}}$ a verdict
$v_j \in \{1, 0.5, 0, \text{n/a}\}$ for match, partial match, mismatch, or
not applicable. The $\text{n/a}$ verdict is used when the corresponding
feature is absent from the original game. The review judge again confirms or overturns each verdict. Visual fidelity is averaged over the
applicable items:
\begin{equation}\text{Score}^{\text{vis}} = \frac{\sum_{j:\, v_j \neq \text{n/a}} v_j}{\left|\{\, j : v_j \neq \text{n/a} \,\}\right|},\end{equation}
so $\text{n/a}$ items are excluded from both the numerator and denominator.

\textbf{Overall score.}
The Overall score of an instance is the weighted
average of implementation consistency, rule consistency, and visual fidelity:
\begin{equation}\begin{aligned}\text{Score} ={}& w_{\text{impl}}\,\text{Score}^{\text{impl}} + w_{\text{rule}}\,\text{Score}^{\text{rule}} \\ &+ w_{\text{vis}}\,\text{Score}^{\text{vis}}, \\ &w_{\text{impl}} + w_{\text{rule}} + w_{\text{vis}} = 1.\end{aligned}\end{equation}
The experiments use equal weights,
$w_{\text{impl}}=w_{\text{rule}}=w_{\text{vis}}=1/3$. Averaging instance-level Overall scores
over a difficulty level, a game, or the entire benchmark yields the aggregate
score at the corresponding granularity.

\section{Experiments}
\label{sec:experiments}

Replication quality depends on both the model and its agent runtime.
The model supplies reasoning, multimodal grounding, context utilization, tool invocation, and instruction following.
The agent runtime provides the control loop, tool wrappers, graphical interface access, file-system state, and trajectory logging.
We compare models under one fixed agent runtime, then compare agent runtimes using the strongest-performing model.

\subsection{Experimental Setup}
\label{sec:experimental-setup}

\textbf{Models.}
We evaluated six models spanning proprietary and open-source families: Claude Opus 4.8 (Anthropic), GPT-5.5 (OpenAI), Grok v4.5 (xAI), GLM-5.3-Flash (Zhipu AI), Kimi K3 (Moonshot AI), and Deepseek-Flash (DeepSeek).
Replication draws on four capabilities that differ across these models: screenshot-based multimodal grounding, long-context utilization, tool-call fidelity, and instruction following under complex constraints.

\textbf{Agent runtimes.}
We evaluated five agent runtimes to examine how their design affects model performance: Claude Code, Codex CLI, Hermes, OpenClaw, and OpenHands.
They differ in the agent loop, tool wrappers, graphical interface access, file-system state management, and trajectory logging.

\textbf{Experimental design.}
We used two controlled comparisons: varying the model under Claude Code, and varying the agent runtime with Claude Opus 4.8.
All configurations shared the same system prompt and maximum time budget $T=2$ hours per run.
Each run used the isolated two-container environment in Section~\ref{sec:gamereplication}, with game access restricted to the HTTP action--screenshot interface.
All configurations followed the same implementation contract (Appendix~\ref{app:example}): a single-file replica whose state-bearing elements are DOM elements, a synchronized global state object, and the screenshot and edit-loop budgets stated in the prompt. Implementation choices were otherwise free, so that the two designs differ only in the manipulated variable.

\textbf{Evaluation models.}
Dedicated LLM judges scored all three evaluation dimensions.
These judges differ from the evaluated models to reduce self-preference bias.
In the cross-model assessment--review protocol, \texttt{gemini-3.6-flash} performs assessment and \texttt{gpt-5.6-sol} independently confirms or overturns its verdicts.

\textbf{Statistical protocol.}
We ran each configuration independently three times per instance.
The instance score is the mean of these three runs.
Values after $\pm$ denote sample standard deviations across instances: $n=125$ across all difficulty levels and $n=25$ within a difficulty level.
The Overall score is the equal-weight average of rule consistency, implementation consistency, and visual fidelity.

\subsection{Model Comparison: Fixed Agent Runtime}

\begin{table*}[t]
  \centering
  \caption{Main results with the agent runtime fixed to Claude Code and the model varied. Cells report mean $\pm$ sample standard deviation across the 125 instances (25 games $\times$ 5 difficulty levels); each instance score averages three independent runs. Rule, Impl, and Visual denote rule consistency, implementation consistency, and visual fidelity, and Overall is their equal-weight mean. Bold marks the highest score or the lowest resource use in each column. Shading encodes the score within each column (darker is higher).}
  \label{tab:main-model}
  \setlength{\tabcolsep}{5pt}
  \renewcommand{\arraystretch}{1.2}
  \fittable{%
\begin{tabular*}{\textwidth}{@{\extracolsep{\fill}}lcccccc}
  \toprule
  \textbf{Model} & \textbf{Rule} & \textbf{Impl} & \textbf{Visual} & \textbf{Overall} & \textbf{Tokens (M)} & \textbf{Time (s)} \\
  \midrule
  Claude Opus 4.8 & \cellcolor[rgb]{0.3,0.66,0.42} $\mathbf{54.2\pm12.0}$& \cellcolor[rgb]{0.32,0.45,0.8} $\mathbf{67.4\pm10.5}$& \cellcolor[rgb]{0.62,0.38,0.66} $\mathbf{93.1\pm5.5}$& \cellcolor[rgb]{0.52,0.52,0.52} $\mathbf{71.6\pm7.4}$& $5.05\pm3.30$ & $1694\pm1159$ \\
  GPT-5.5         & \cellcolor[rgb]{0.62,0.816,0.685} $26.6\pm16.2$& \cellcolor[rgb]{0.583,0.662,0.877} $39.4\pm19.3$& \cellcolor[rgb]{0.738,0.572,0.766} $62.8\pm21.3$& \cellcolor[rgb]{0.699,0.699,0.699} $43.0\pm15.6$& $\mathbf{2.33\pm1.99}$ & $\mathbf{1355\pm358}$ \\
  Grok v4.5       & \cellcolor[rgb]{0.743,0.875,0.787} $16.0\pm16.5$& \cellcolor[rgb]{0.74,0.79,0.924} $22.6\pm21.1$& \cellcolor[rgb]{0.807,0.686,0.828} $45.0\pm26.2$& \cellcolor[rgb]{0.793,0.793,0.793} $27.9\pm18.7$& $12.19\pm7.74$ & $1723\pm685$ \\
  GLM-5.3-Flash         & \cellcolor[rgb]{0.829,0.917,0.858} $8.6\pm13.2$& \cellcolor[rgb]{0.825,0.858,0.948} $13.6\pm16.9$& \cellcolor[rgb]{0.861,0.773,0.875} $31.3\pm24.6$& \cellcolor[rgb]{0.856,0.856,0.856} $17.8\pm15.4$& $17.12\pm10.62$ & $3937\pm1390$ \\
  Kimi K3         & \cellcolor[rgb]{0.88,0.942,0.901} $4.2\pm7.9$& \cellcolor[rgb]{0.881,0.904,0.965} $7.6\pm13.9$& \cellcolor[rgb]{0.88,0.804,0.892} $26.4\pm25.4$& \cellcolor[rgb]{0.888,0.888,0.888} $12.8\pm13.9$& $20.98\pm15.71$ & $1982\pm1010$ \\
  Deepseek-Flash & \cellcolor[rgb]{0.916,0.959,0.93} $1.1\pm3.8$& \cellcolor[rgb]{0.918,0.934,0.976} $3.6\pm6.5$& \cellcolor[rgb]{0.954,0.926,0.959} $7.2\pm12.8$& \cellcolor[rgb]{0.942,0.942,0.942} $4.0\pm6.4$& $16.42\pm5.94$ & $2912\pm614$ \\
  \bottomrule
  \end{tabular*}%
}
\end{table*}

Under the fixed Claude Code agent runtime, model choice changes replication quality while the orchestration logic remains identical.
Table~\ref{tab:main-model} supports four findings.

\textbf{Finding 1: Under the same agent runtime, the six models differ widely in implementation consistency.}
In Table~\ref{tab:main-model}, implementation consistency ranges from 67.4 for Claude Opus 4.8 to 3.6 for Deepseek-Flash, a gap of 63.8 points; their visual fidelity scores are 93.1 and 7.2.
GPT-5.5 follows Claude Opus 4.8 at 39.4, and the remaining four models score below 25.
Replication quality under the same agent runtime therefore varies widely with the model.

\textbf{Finding 2: Token consumption is negatively associated with Overall score across the evaluated models.}
GLM-5.3-Flash, Kimi K3, and Deepseek-Flash have the lowest Overall scores in Table~\ref{tab:main-model}.
Their token consumption (17.12\,M, 20.98\,M, and 16.42\,M) is three to four times that of Claude Opus 4.8 (5.05\,M).
Across model means, token consumption is negatively associated with Overall score (Spearman $\rho=-0.77$, $n=6$).
Trajectory analysis (Section~\ref{sec:failure-submode}) identifies repeated trial and error and budget exhaustion among weaker models.
This association is descriptive: it is computed over six model means and does not test whether a larger budget would change the score of a given model.

\textbf{Finding 3: Model rankings agree across dimensions, and visual fidelity exceeds implementation consistency, which exceeds rule consistency.}
All dimensions in Table~\ref{tab:main-model} rank the models as Claude Opus 4.8 $>$ GPT-5.5 $>$ Grok v4.5 $>$ GLM-5.3-Flash $>$ Kimi K3 $>$ Deepseek-Flash.
Thus, the ranking does not depend on a single dimension.
Under the current rubrics, every model also scores highest on visual fidelity and lowest on rule consistency, with implementation consistency in between.

\textbf{Finding 4: Stronger models show smaller relative Overall score declines as difficulty increases.}
Table~\ref{tab:decay-model-overall} shows that Claude Opus 4.8 declines from 72.1 to 68.6 from L1 to L5 (3.5 points, or 4.9\%).
GPT-5.5 also shows a limited, non-monotonic decline (46.6$\to$42.2, 9.4\%).
Grok v4.5 declines from 35.9 to 15.5 (56.8\%), and GLM-5.3-Flash from 25.2 to 11.0 (56.3\%).
Kimi K3 and Deepseek-Flash decline by 5.1 and 3.4 points, respectively.
Their scores approach the lower bound, limiting absolute changes, but their relative declines remain 33.1\% and 63.0\%.
For weaker models, per-level standard deviations often match or exceed their means.
At L1, Kimi K3 scores 11.9$\pm$21.9 on implementation consistency, and GLM-5.3-Flash scores 20.7$\pm$21.9.
These statistics are consistent with highly uneven performance, with low scores on many instances and higher scores on a few.
Stronger models thus combine higher means with smaller across-instance dispersion at the harder levels; the reported standard deviations describe variation across instances, not across repeated runs.
Appendix~\ref{app:difficulty-decay} reports all dimensions by difficulty.

\begin{table*}[t]
  \centering
  \caption{Replication scores by difficulty level (L1 $\to$ L5, Overall score) with the model varied and the agent runtime fixed (Claude Code). Each cell reports mean $\pm$ sample standard deviation across the 25 instances at that level, after averaging three independent runs per instance; \textbf{All} pools all 125 instances. Shading encodes the score within each row (darker is higher).}
  \label{tab:decay-model-overall}
  \setlength{\tabcolsep}{5pt}
  \renewcommand{\arraystretch}{1.2}
  \fittable{%
\begin{tabular*}{\textwidth}{@{\extracolsep{\fill}}lcccccc}
  \toprule
  \textbf{Model} & \textbf{L1} & \textbf{L2} & \textbf{L3} & \textbf{L4} & \textbf{L5} & \textbf{All} \\
  \midrule
  Claude Opus 4.8 & \cellcolor[rgb]{0.516,0.609,0.858} $72.1\pm8.0$& \cellcolor[rgb]{0.32,0.45,0.8} $74.0\pm4.3$& \cellcolor[rgb]{0.423,0.534,0.83} $73.0\pm5.2$& \cellcolor[rgb]{0.712,0.767,0.915} $70.2\pm8.1$& \cellcolor[rgb]{0.878,0.901,0.964} $68.6\pm9.6$& $71.6\pm7.4$ \\
  GPT-5.5         & \cellcolor[rgb]{0.395,0.706,0.499} $46.6\pm11.6$& \cellcolor[rgb]{0.3,0.66,0.42} $48.5\pm14.9$& \cellcolor[rgb]{0.699,0.854,0.751} $40.5\pm13.3$& \cellcolor[rgb]{0.874,0.939,0.896} $37.0\pm15.6$& \cellcolor[rgb]{0.614,0.813,0.681} $42.2\pm19.6$& $43.0\pm15.6$ \\
  Grok v4.5       & \cellcolor[rgb]{0.92,0.62,0.28} $35.9\pm15.3$& \cellcolor[rgb]{0.937,0.699,0.43} $30.7\pm21.0$& \cellcolor[rgb]{0.944,0.733,0.494} $28.5\pm19.9$& \cellcolor[rgb]{0.943,0.728,0.485} $28.8\pm17.6$& \cellcolor[rgb]{0.986,0.932,0.87} $15.5\pm13.8$& $27.9\pm18.7$ \\
  GLM-5.3-Flash         & \cellcolor[rgb]{0.62,0.4,0.7} $25.2\pm18.1$& \cellcolor[rgb]{0.646,0.442,0.721} $24.0\pm16.2$& \cellcolor[rgb]{0.817,0.712,0.856} $16.2\pm11.8$& \cellcolor[rgb]{0.892,0.83,0.915} $12.8\pm11.6$& \cellcolor[rgb]{0.932,0.892,0.946} $11.0\pm13.6$& $17.8\pm15.4$ \\
  Kimi K3         & \cellcolor[rgb]{0.28,0.66,0.68} $15.4\pm17.2$& \cellcolor[rgb]{0.326,0.682,0.701} $15.0\pm14.7$& \cellcolor[rgb]{0.639,0.829,0.839} $12.3\pm8.9$& \cellcolor[rgb]{0.801,0.906,0.912} $10.9\pm14.4$& \cellcolor[rgb]{0.87,0.939,0.942} $10.3\pm13.1$& $12.8\pm13.9$ \\
  Deepseek-Flash & \cellcolor[rgb]{0.85,0.4,0.42} $5.4\pm9.3$& \cellcolor[rgb]{0.893,0.574,0.588} $4.2\pm5.1$& \cellcolor[rgb]{0.897,0.588,0.602} $4.1\pm6.6$& \cellcolor[rgb]{0.901,0.603,0.616} $4.0\pm5.2$& \cellcolor[rgb]{0.973,0.892,0.896} $2.0\pm4.4$& $4.0\pm6.4$ \\
  \bottomrule
  \end{tabular*}%
}
\end{table*}

\subsection{Agent Runtime Comparison: Fixed Model}

\begin{table*}[t]
  \centering
  \caption{Main results with the model fixed to Claude Opus 4.8 and the agent runtime varied. Cells report mean $\pm$ sample standard deviation across the 125 instances (25 games $\times$ 5 difficulty levels); each instance score averages three independent runs. Rule, Impl, and Visual denote rule consistency, implementation consistency, and visual fidelity, and Overall is their equal-weight mean. Bold marks the highest score or the lowest resource use in each column. Shading encodes the score within each column (darker is higher).}
  \label{tab:main-harness}
  \setlength{\tabcolsep}{5pt}
  \renewcommand{\arraystretch}{1.2}
  \fittable{%
\begin{tabular*}{\textwidth}{@{\extracolsep{\fill}}lcccccc}
  \toprule
  \textbf{Runtime} & \textbf{Rule} & \textbf{Impl} & \textbf{Visual} & \textbf{Overall} & \textbf{Tokens (M)} & \textbf{Time (s)} \\
  \midrule
  Claude Code & \cellcolor[rgb]{0.3,0.66,0.42} $\mathbf{54.2\pm12.0}$& \cellcolor[rgb]{0.32,0.45,0.8} $\mathbf{67.4\pm10.5}$& \cellcolor[rgb]{0.62,0.38,0.66} $\mathbf{93.1\pm5.5}$& \cellcolor[rgb]{0.52,0.52,0.52} $\mathbf{71.6\pm7.4}$& $5.05\pm3.30$ & $\mathbf{1694\pm1159}$ \\
  Codex       & \cellcolor[rgb]{0.527,0.77,0.608} $49.7\pm12.7$& \cellcolor[rgb]{0.456,0.56,0.84} $64.0\pm11.3$& \cellcolor[rgb]{0.669,0.46,0.704} $91.2\pm7.4$& \cellcolor[rgb]{0.624,0.624,0.624} $68.3\pm7.9$& $\mathbf{4.18\pm2.20}$ & $2419\pm1211$ \\
  Hermes      & \cellcolor[rgb]{0.547,0.78,0.625} $49.3\pm13.5$& \cellcolor[rgb]{0.504,0.598,0.854} $62.8\pm13.5$& \cellcolor[rgb]{0.705,0.518,0.736} $89.8\pm8.3$& \cellcolor[rgb]{0.656,0.656,0.656} $67.3\pm9.1$& $6.69\pm3.87$ & $2268\pm1106$ \\
  OpenClaw    & \cellcolor[rgb]{0.916,0.959,0.93} $42.0\pm18.8$& \cellcolor[rgb]{0.815,0.85,0.945} $55.0\pm21.5$& \cellcolor[rgb]{0.864,0.779,0.879} $83.6\pm28.3$& \cellcolor[rgb]{0.879,0.879,0.879} $60.2\pm21.4$& $8.16\pm4.77$ & $2875\pm1538$ \\
  OpenHands   & \cellcolor[rgb]{0.906,0.954,0.922} $42.2\pm20.0$& \cellcolor[rgb]{0.918,0.934,0.976} $52.4\pm22.5$& \cellcolor[rgb]{0.954,0.926,0.959} $80.1\pm20.6$& \cellcolor[rgb]{0.942,0.942,0.942} $58.2\pm19.5$& $13.88\pm7.85$ & $3018\pm1353$ \\
  \bottomrule
  \end{tabular*}%
}
\end{table*}

\textbf{Finding 5: Agent runtimes yield different replication scores for the same model.}
In Table~\ref{tab:main-harness}, implementation consistency follows Claude Code 67.4 $>$ Codex 64.0 $>$ Hermes 62.8 $>$ OpenClaw 55.0 $>$ OpenHands 52.4.
The highest and lowest scores differ by 15.0 points.
Visual fidelity follows the same order (93.1$\to$80.1).
Rule consistency preserves the top-three order, while OpenHands (42.2) and OpenClaw (42.0) are nearly tied.
Agent runtime design therefore affects how effectively the same model performs replication, including through multi-turn orchestration and tool invocation.

\textbf{Finding 6: Implementation consistency varies more across models than across agent runtimes.}
The model comparison spans 63.8 points in implementation consistency (Claude Opus 4.8: 67.4; Deepseek-Flash: 3.6).
The agent runtime comparison spans 15.0 points (67.4 versus 52.4).
Within these comparisons, model choice accounts for greater variation in replication quality than agent runtime choice.

\textbf{Finding 7: Lower-scoring agent runtimes show greater variability on difficult instances.}
Table~\ref{tab:decay-harness-overall} shows relatively stable Overall scores for Claude Code and Codex (e.g., Codex: 70.2$\to$65.9).
OpenClaw declines from 70.6 at L1 to 53.5 at L5, a 17.1-point decrease, with across-instance standard deviations exceeding 19 from L2 onward.
Its reported dispersion is $\pm$25.0 for the L5 Overall score and $\pm$35.2 for L3 visual fidelity.
This dispersion indicates uneven outcomes across instances on harder levels.
OpenHands starts lower (62.3 at L1) and shows large variability across levels.
Claude Code and Codex maintain lower variability across difficulty levels.
Agent runtimes thus differ in both mean quality and across-instance dispersion on difficult instances.
Appendix~\ref{app:difficulty-decay} reports the full results for each dimension.

\begin{table*}[t]
  \centering
  \caption{Replication scores by difficulty level (L1 $\to$ L5, Overall score) with the model fixed (Claude Opus 4.8) and the agent runtime varied. Each cell reports mean $\pm$ sample standard deviation across the 25 instances at that level, after averaging three independent runs per instance; \textbf{All} pools all 125 instances. Shading encodes the score within each row (darker is higher).}
  \label{tab:decay-harness-overall}
  \setlength{\tabcolsep}{5pt}
  \renewcommand{\arraystretch}{1.2}
  \fittable{%
\begin{tabular*}{\textwidth}{@{\extracolsep{\fill}}lcccccc}
  \toprule
  \textbf{Runtime} & \textbf{L1} & \textbf{L2} & \textbf{L3} & \textbf{L4} & \textbf{L5} & \textbf{All} \\
  \midrule
  Claude Code & \cellcolor[rgb]{0.516,0.609,0.858} $72.1\pm8.0$& \cellcolor[rgb]{0.32,0.45,0.8} $74.0\pm4.3$& \cellcolor[rgb]{0.423,0.534,0.83} $73.0\pm5.2$& \cellcolor[rgb]{0.712,0.767,0.915} $70.2\pm8.1$& \cellcolor[rgb]{0.878,0.901,0.964} $68.6\pm9.6$& $71.6\pm7.4$ \\
  Codex       & \cellcolor[rgb]{0.3,0.66,0.42} $70.2\pm6.6$& \cellcolor[rgb]{0.327,0.673,0.442} $70.0\pm6.1$& \cellcolor[rgb]{0.393,0.705,0.497} $69.5\pm7.7$& \cellcolor[rgb]{0.874,0.939,0.896} $65.9\pm9.0$& \cellcolor[rgb]{0.874,0.939,0.896} $65.9\pm9.0$& $68.3\pm7.9$ \\
  Hermes      & \cellcolor[rgb]{0.928,0.656,0.348} $70.2\pm6.2$& \cellcolor[rgb]{0.92,0.62,0.28} $71.2\pm7.3$& \cellcolor[rgb]{0.946,0.745,0.518} $67.7\pm8.8$& \cellcolor[rgb]{0.967,0.842,0.701} $65.0\pm9.2$& \cellcolor[rgb]{0.986,0.932,0.87} $62.5\pm11.1$& $67.3\pm9.1$ \\
  OpenClaw    & \cellcolor[rgb]{0.62,0.4,0.7} $70.6\pm7.5$& \cellcolor[rgb]{0.742,0.593,0.796} $63.9\pm20.3$& \cellcolor[rgb]{0.882,0.814,0.907} $56.2\pm26.4$& \cellcolor[rgb]{0.871,0.797,0.899} $56.8\pm19.6$& \cellcolor[rgb]{0.932,0.892,0.946} $53.5\pm25.0$& $60.2\pm21.4$ \\
  OpenHands   & \cellcolor[rgb]{0.433,0.732,0.748} $62.3\pm14.8$& \cellcolor[rgb]{0.28,0.66,0.68} $65.6\pm12.4$& \cellcolor[rgb]{0.703,0.86,0.868} $56.5\pm25.1$& \cellcolor[rgb]{0.824,0.917,0.922} $53.9\pm19.3$& \cellcolor[rgb]{0.87,0.939,0.942} $52.9\pm21.8$& $58.2\pm19.5$ \\
  \bottomrule
  \end{tabular*}%
}
\end{table*}

\subsection{Process Behavior Analysis}
\label{sec:process-metrics}

Process metrics characterize how models and agent runtimes acquire visual evidence and interact with the environment.
We measure three aspects of each execution trajectory:
\begin{itemize}
  \item \textbf{Visual-observation volume}: the number of screenshots of the original game and replica, measuring the volume of visual feedback used for learning and self-checking.
  \item \textbf{Environment-interaction volume}: the total number and distribution of HTTP calls to the game-server endpoints, including screenshot requests, measuring interaction intensity and interface use.
  \item \textbf{Tool-call count}: the total number of tool invocations within the agent runtime.
\end{itemize}
For these metrics, we distinguish the exploration phase (observing the original game) from the self-check phase (checking the replica).
We report means across instances for both controlled comparisons: varying the model under a fixed agent runtime, and varying the agent runtime with a fixed model. Visual-observation volume is reported for the model comparison only.

\subsubsection{Model Comparison}

\begin{figure*}[t]
  \centering
  \begin{subfigure}[t]{0.48\textwidth}
    \centering
    \includegraphics[width=\linewidth,height=\processfigheight,keepaspectratio]{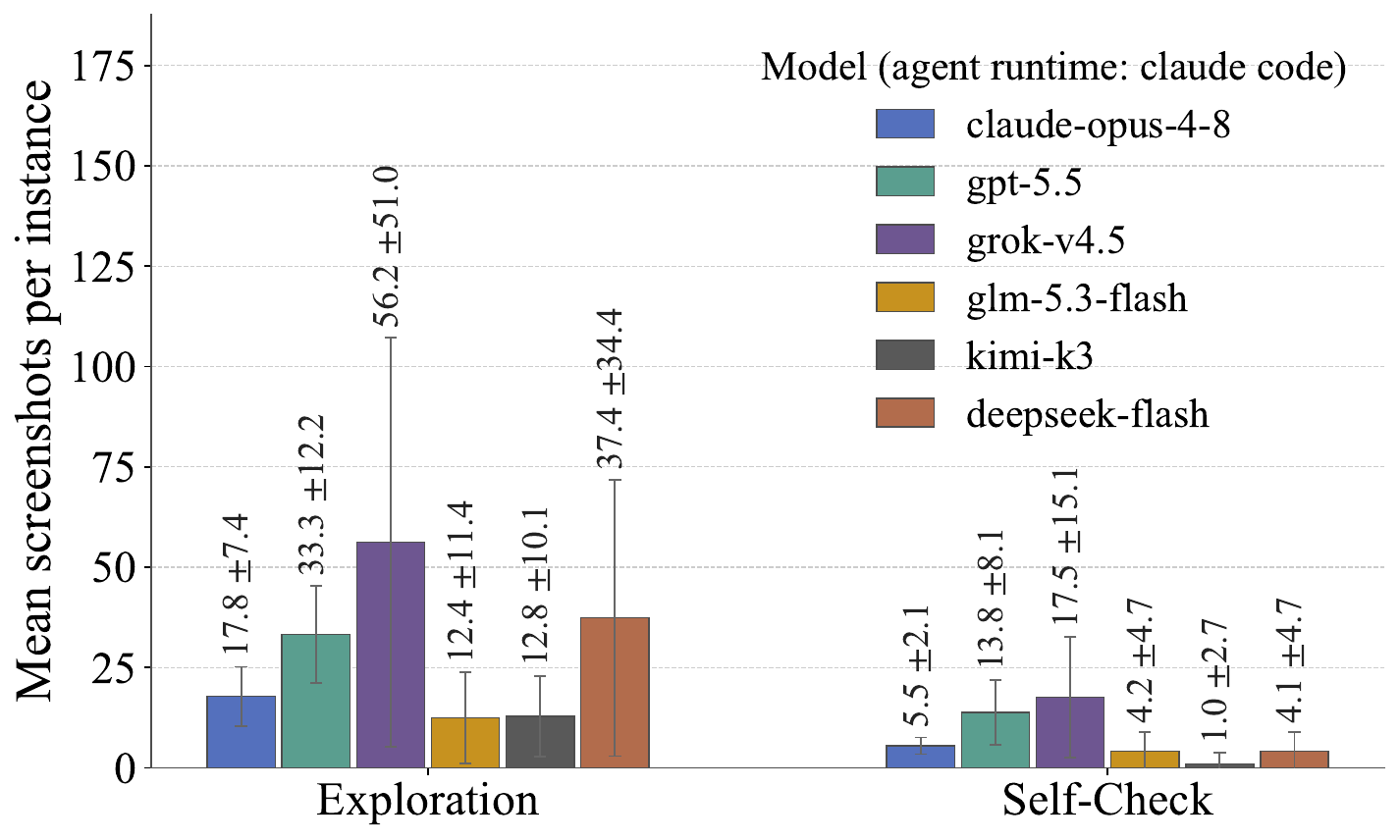}
    \caption{Mean screenshots per instance in the exploration and self-check phases.}
    \label{fig:dm-shots-a}
  \end{subfigure}
  \begin{subfigure}[t]{0.48\textwidth}
    \centering
    \includegraphics[width=\linewidth,height=\processfigheight,keepaspectratio]{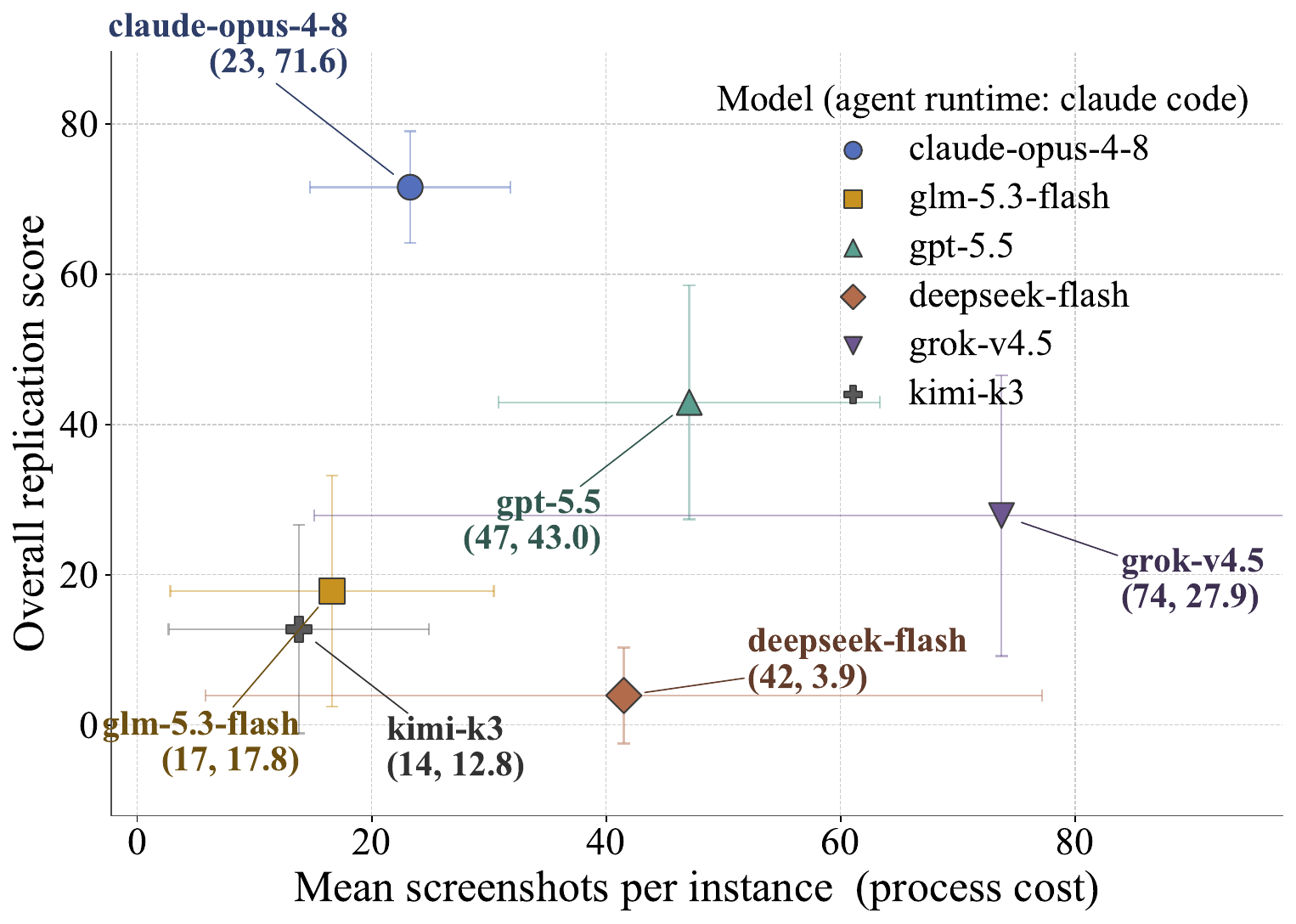}
    \caption{Total screenshots per instance versus Overall score.}
    \label{fig:dm-shots-b}
  \end{subfigure}
  \caption{Visual observation across models with the agent runtime fixed to Claude Code. (a) Mean screenshots in the exploration and self-check phases. (b) Total screenshots across both phases versus Overall score.}
  \label{fig:dm-shots}
\end{figure*}

\textbf{Visual-observation volume.}
Screenshot counts measure how much an agent observes, rather than how much it understands. Figure~\ref{fig:dm-shots}(a) shows that all six models take far fewer screenshots during self-check than during exploration (Claude Opus 4.8: 5.5 versus 17.8; Grok v4.5: 17.5 versus 56.2). Figure~\ref{fig:dm-shots}(b) shows that total screenshots and Overall score are not monotonically related. Claude Opus 4.8 achieves the highest Overall score (71.6) with 23 screenshots, less than one-third of the largest count. Grok v4.5 takes the most screenshots (74) but scores only 27.9; Deepseek-Flash takes 42 screenshots but has the lowest Overall score (4.0). Screenshot counts therefore do not measure effective observation or model capability.

\begin{figure*}[t]
  \centering
  \begin{subfigure}[t]{0.46\textwidth}
    \includegraphics[width=\linewidth,height=\processfigheight,keepaspectratio]{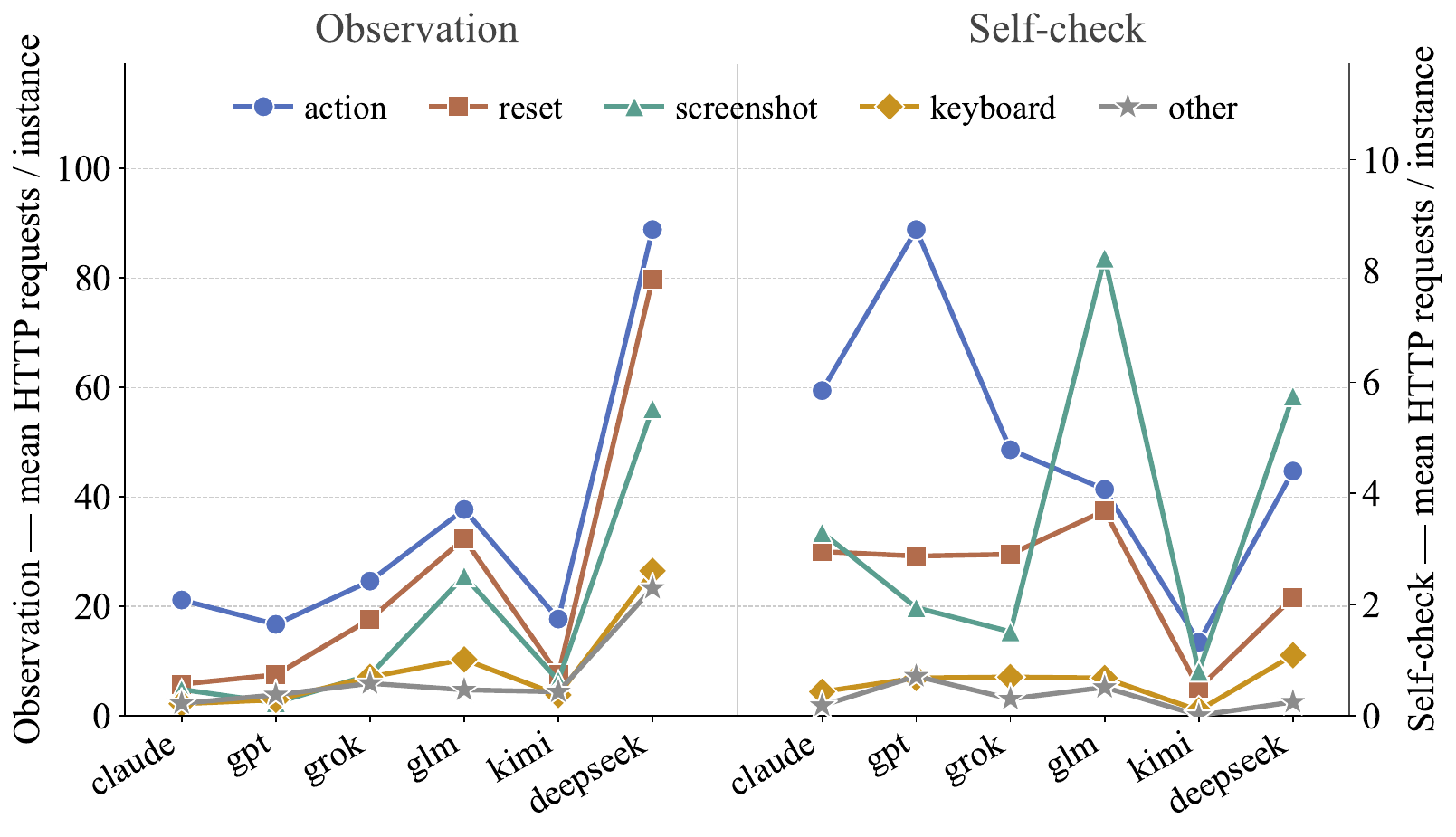}
    \caption{Mean HTTP calls per instance per endpoint.}
    \label{fig:dm-env-a}
  \end{subfigure}\hfill
  \begin{subfigure}[t]{0.46\textwidth}
    \includegraphics[width=\linewidth,height=\processfigheight,keepaspectratio]{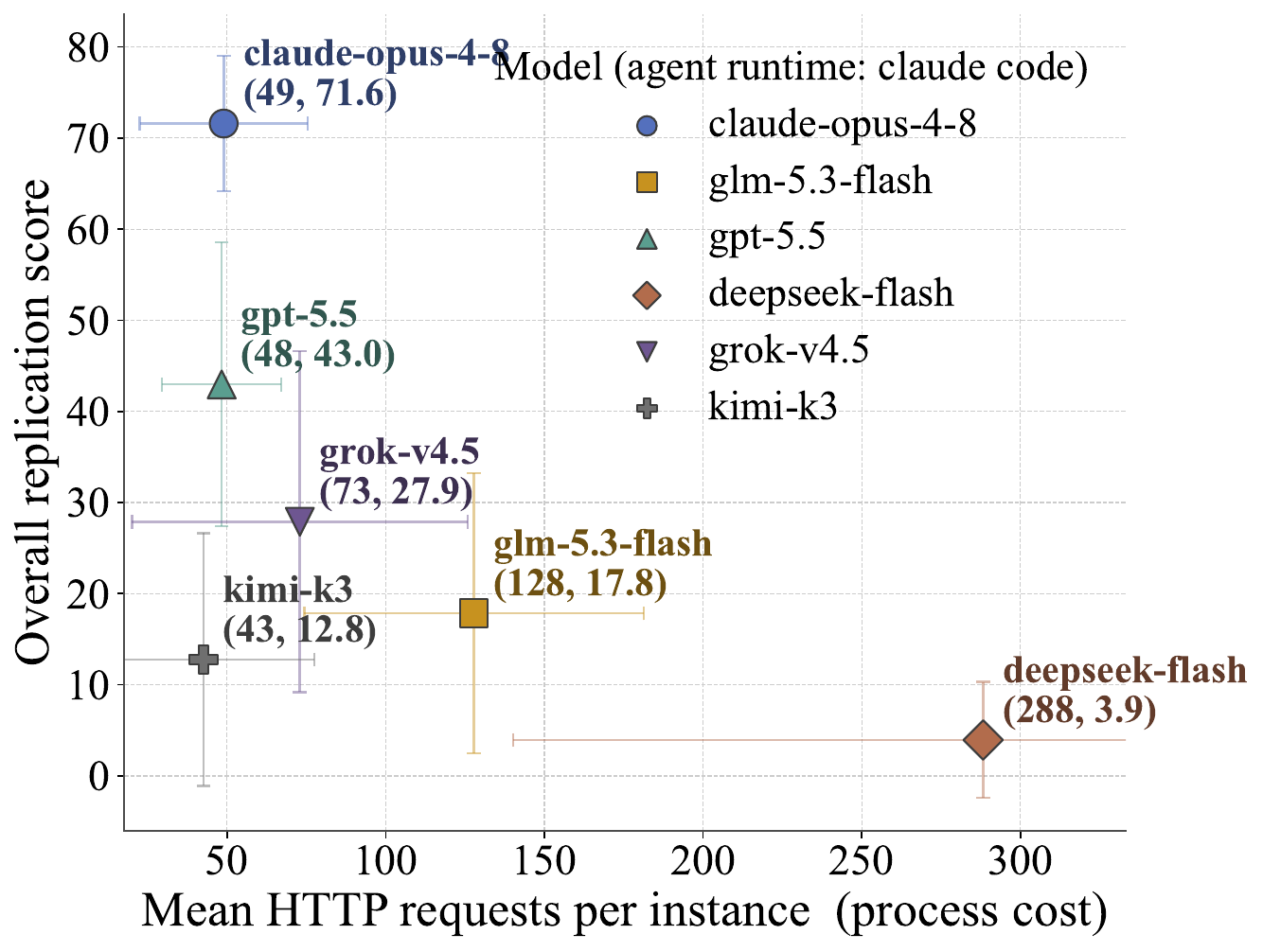}
    \caption{Total HTTP calls versus Overall score.}
    \label{fig:dm-env-b}
  \end{subfigure}
  \caption{Environment interaction across models (agent runtime fixed to Claude Code).
  (a) Mean per-instance HTTP calls broken down over five endpoint types (action / reset / screenshot / keyboard / other), separating the exploration phase (left axis) from the self-check phase (right axis).
  (b) Total HTTP calls per instance (process cost) plotted against Overall score.}
  \label{fig:dm-env}
\end{figure*}

\textbf{Environment-interaction volume.}
Frequent HTTP interaction is associated with repeated trial and error rather than consistently higher replication quality.
Figure~\ref{fig:dm-env}(a) separates exploration-phase calls from self-check-phase calls across five endpoint types: action, reset, screenshot, keyboard, and other.
The \emph{other} category combines scroll, drag, and hover.
Action calls dominate exploration for every model.
Deepseek-Flash and GLM-5.3-Flash also make more reset and screenshot calls; Deepseek-Flash makes roughly 23 other calls per instance, mostly drags.
This pattern reflects repeated resets, screenshots, and dragging.
Self-check calls are far fewer than exploration calls: the self-check axis spans about one-tenth of the exploration axis.
Figure~\ref{fig:dm-env}(b) shows no monotonic relationship between total HTTP calls and Overall score.
Claude Opus 4.8 scores 71.6 with 49 calls, less than one-fifth of the count for Deepseek-Flash.
Deepseek-Flash makes 288 calls but scores 4.0; GLM-5.3-Flash makes 128 calls and scores 17.8.
Call volume alone therefore does not indicate replication capability.

\begin{figure*}[t]
  \centering
  \begin{subfigure}[t]{0.46\textwidth}
    \centering
    \includegraphics[width=\linewidth,height=3.9cm,keepaspectratio]{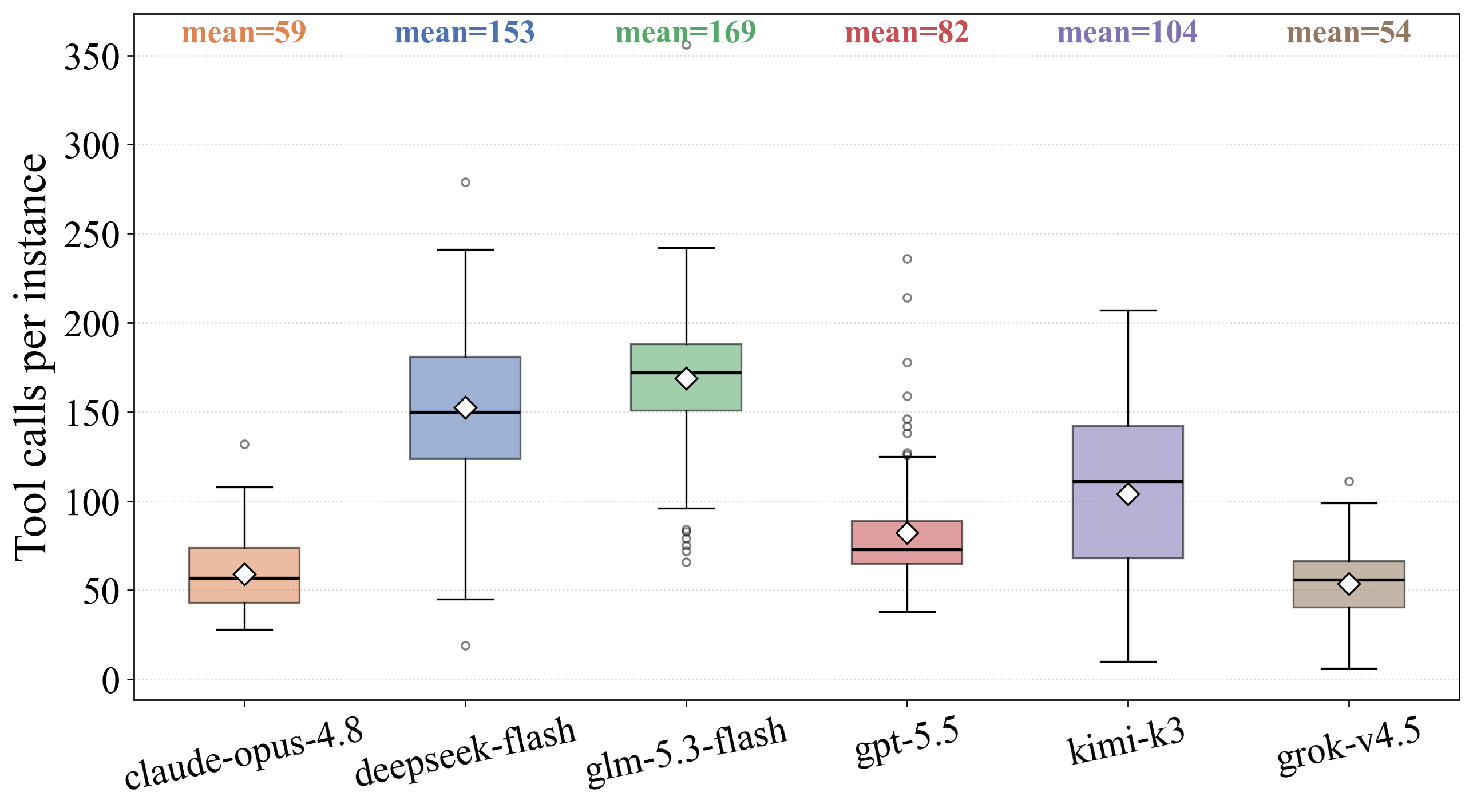}
    \caption{Across models (agent runtime fixed to Claude Code).}
    \label{fig:dm-tools}
  \end{subfigure}\hfill
  \begin{subfigure}[t]{0.46\textwidth}
    \centering
    \includegraphics[width=\linewidth,height=3.9cm,keepaspectratio]{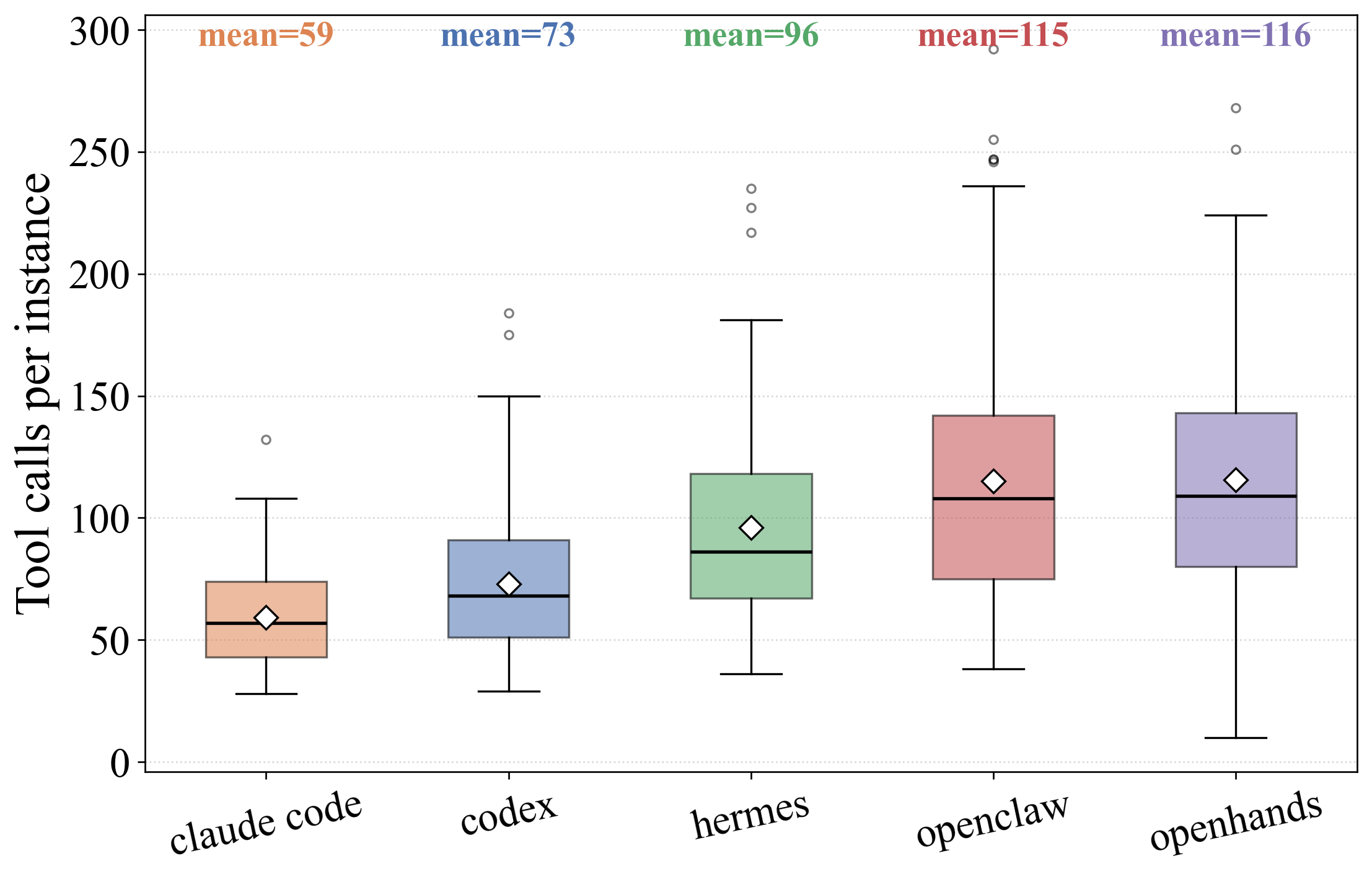}
    \caption{Across agent runtimes (model fixed to Claude Opus 4.8).}
    \label{fig:dh-tools}
  \end{subfigure}
  \caption{Tool-call counts per instance. Each box plot shows the per-instance distribution of tool calls; the white diamond marks the mean $\mu$, annotated above the box. \textbf{(a)} varying the model with the agent runtime fixed; \textbf{(b)} varying the agent runtime with the model fixed.}
  \label{fig:tools}
\end{figure*}

\textbf{Tool-call count.}
Under the fixed agent runtime, models differ by about 3$\times$ in mean tool-call count, without a corresponding ordering in quality.
Figure~\ref{fig:dm-tools} shows means of 54 for Grok v4.5, 59 for Claude Opus 4.8, 82 for GPT-5.5, 104 for Kimi K3, 153 for Deepseek-Flash, and 169 for GLM-5.3-Flash.
Grok v4.5 and Claude Opus 4.8 make similar numbers of calls, but their Overall scores differ by 43.7 points.
Deepseek-Flash and GLM-5.3-Flash make nearly three times as many calls, consistent with repeated probing before convergence.
With the agent runtime held constant, these differences reflect model-dependent interaction strategies.

\subsubsection{Agent Runtime Comparison}

\begin{figure*}[t]
  \centering
  \begin{subfigure}[t]{0.46\textwidth}
    \includegraphics[width=\linewidth,height=\processfigheight,keepaspectratio]{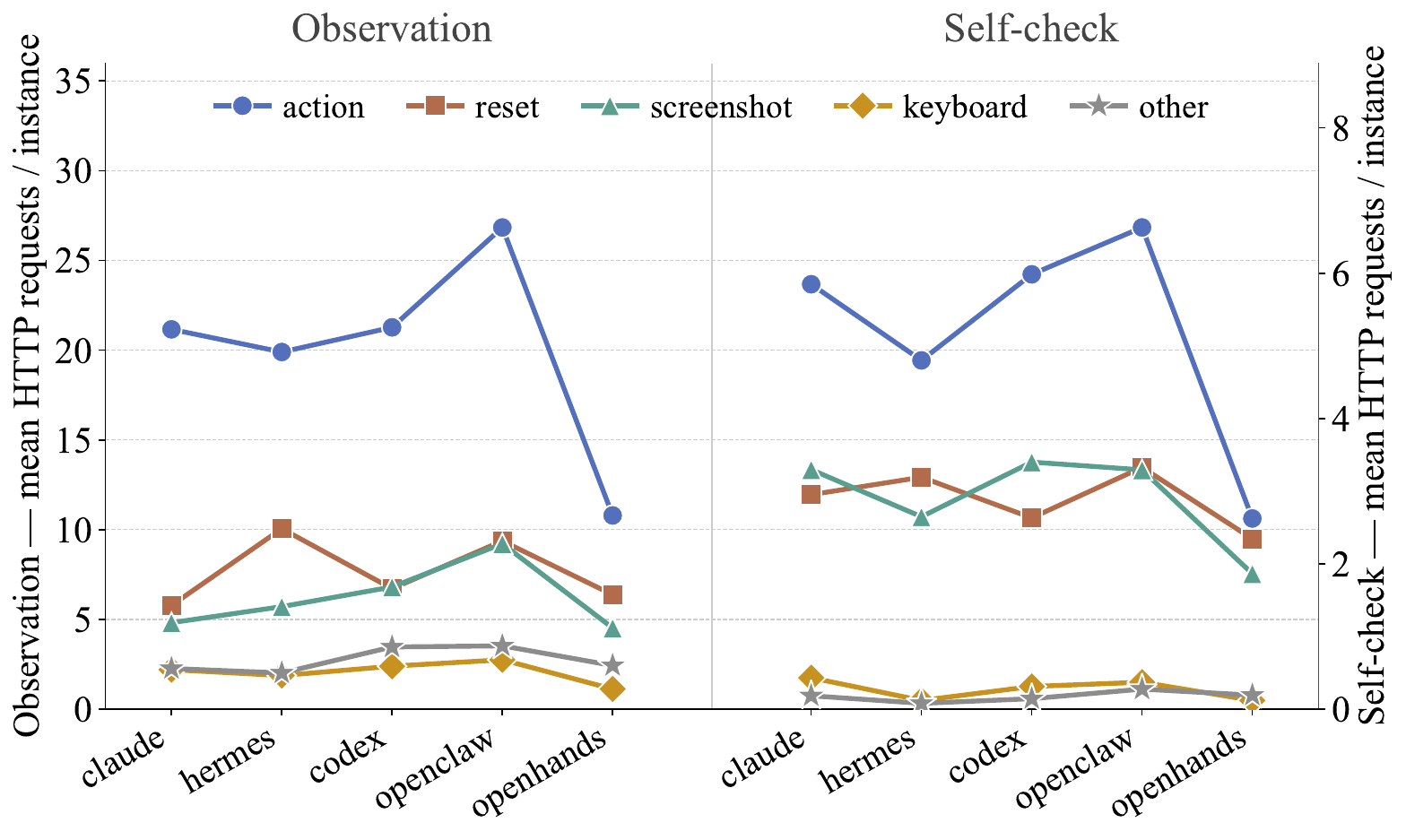}
    \caption{Mean HTTP calls per instance per endpoint.}
    \label{fig:dh-env-a}
  \end{subfigure}
  \begin{subfigure}[t]{0.46\textwidth}
    \includegraphics[width=\linewidth,height=\processfigheight,keepaspectratio]{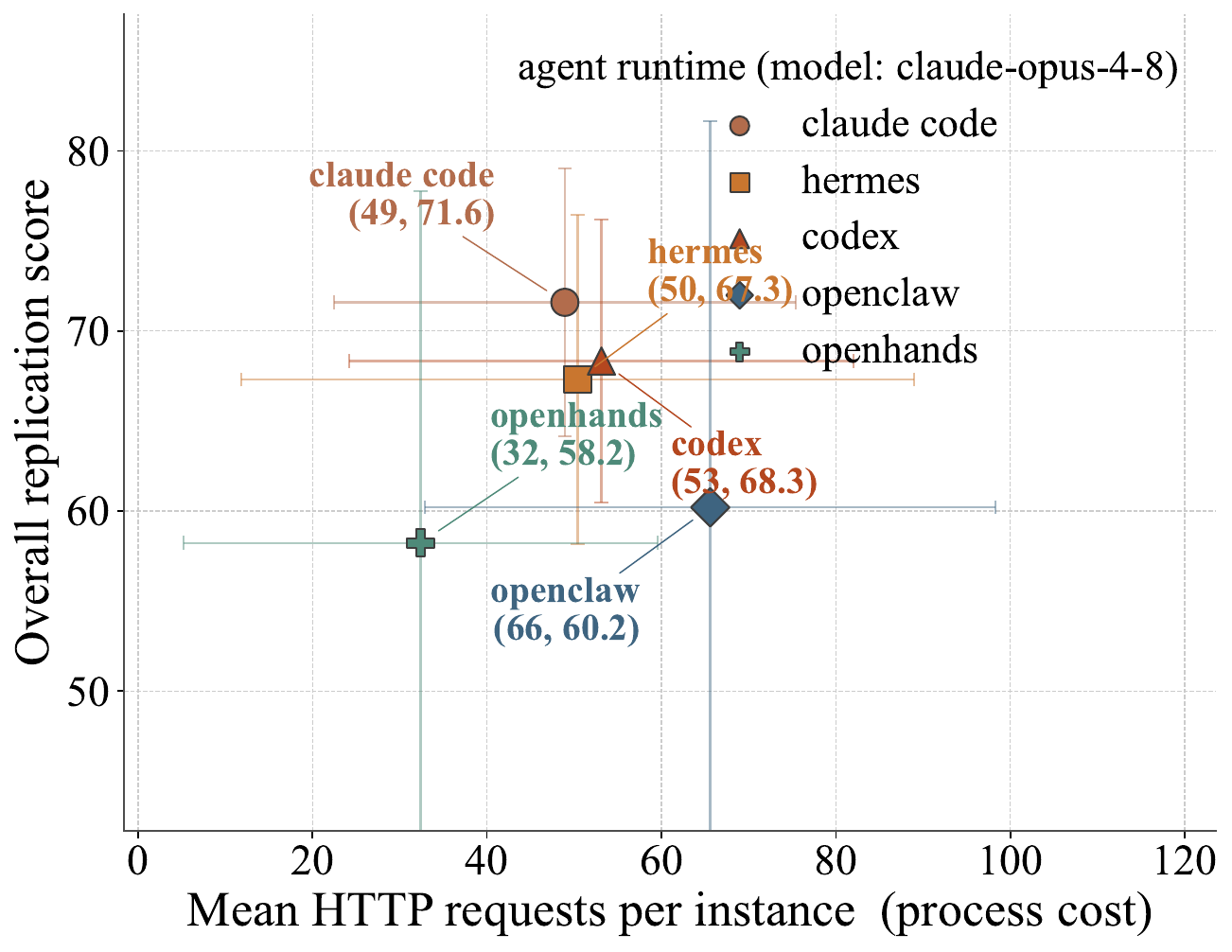}
    \caption{Total HTTP calls versus Overall score.}
    \label{fig:dh-env-b}
  \end{subfigure}
  \caption{Environment interaction across agent runtimes (model fixed to Claude Opus 4.8).
  (a) Mean per-instance HTTP calls over the five endpoint types, separating the exploration phase (left axis) from the self-check phase (right axis).
  (b) Total HTTP calls per instance (process cost) plotted against Overall score.}
  \label{fig:dh-env}
\end{figure*}

\textbf{Environment-interaction volume.}
Varying the agent runtime with Claude Opus 4.8 produces less variation in HTTP interaction and replication quality than varying the model.
Figure~\ref{fig:dh-env}(a) shows similar endpoint compositions across the five agent runtimes.
Action calls dominate exploration, followed by reset and screenshot calls.
Self-check calls are again fewer: the self-check axis spans about one-fourth of the exploration axis.
OpenClaw has the highest overall endpoint-call volume and OpenHands the lowest, while their relative compositions remain similar.
Figure~\ref{fig:dh-env}(b) places the agent runtimes in a compact range: roughly 32--66 calls and Overall scores of 58--72.
Claude Code scores 71.6 with 49 calls; OpenClaw makes the most calls (66) but scores 60.2.
OpenHands makes the fewest calls (32) and has the lowest score (58.2).
Compared with the model comparison (Figure~\ref{fig:dm-env}), the agent runtimes differ less in call volume, while their Overall scores still span 58.2--71.6.

\textbf{Tool-call count.}
Changing the agent runtime with the model fixed changes mean tool-call counts by about 2$\times$.
Figure~\ref{fig:dh-tools} shows an increase from Claude Code (59) to Codex (73), Hermes (96), OpenClaw (115), and OpenHands (116).
Because the model is fixed, the difference is attributable to agent runtime design, including tool granularity and the control loop.
Agent runtime choice therefore changes the number of tool interactions required by the same model.

\subsection{Failure Analysis}
\label{sec:failure-analysis}
\subsubsection{Outcome-level Failure Decomposition}
\label{sec:failure-decomp}

An ordered outcome taxonomy identifies the first unmet requirement for each replica.
It comprises four failure tiers and one success tier (Table~\ref{tab:failure-tiers} in Appendix~\ref{app:failure-tables}).
The gates are: loadable page $\to$ recognizable game $\to$ correct rule implementation $\to$ visual match.
Each instance is assigned to its first unmet gate; replicas that pass all gates are classified as Faithful.
The mechanism score is $S^{\text{mech}}=(\text{Score}^{\text{impl}}+\text{Score}^{\text{rule}})/2$, with all scores on a 0--100 scale.
The mutually exclusive, exhaustive categories sum to 100\% for each configuration, enabling comparisons of failure composition across models and agent runtimes.

\begin{figure*}[t]
  \centering
  \includegraphics[width=\textwidth]{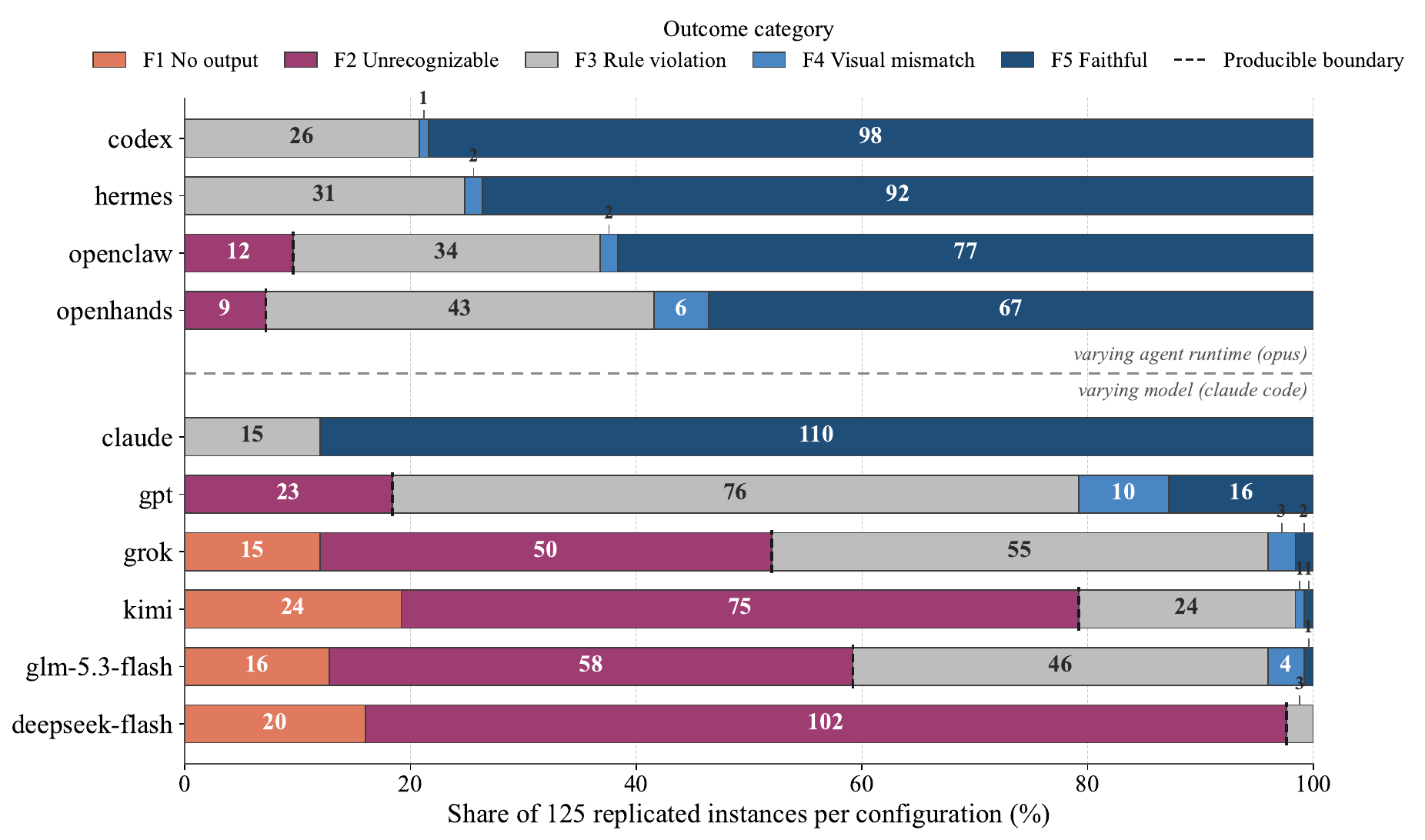}
  \caption{Outcome-level failure decomposition of game replication. Each bar shows a configuration's share of the 125 instances across the five tiers (F1 No output, F2 Unrecognizable, F3 Rule violation, F4 Visual mismatch, F5 Faithful). The dashed line separates missing or unrecognizable replicas (F1--F2) from recognizable ones (F3--F5).}
  \label{fig:failure-decomp}
\end{figure*}

\textbf{Outcome distributions vary more across models than across agent runtimes.}
Figure~\ref{fig:failure-decomp} shows that, under Claude Code, the Faithful share ranges from 88\% for Claude Opus 4.8 to 0.8\% for GLM-5.3-Flash and Kimi K3 and zero for Deepseek-Flash.
With Claude Opus 4.8 fixed, agent runtimes retain Faithful shares of 54\%--88\%.
Failure types also vary with model capability.
Claude Code, Codex, and Hermes produce no No output or Unrecognizable instances; their remaining errors concentrate in Rule violation.
These replicas are recognizable but deviate in gameplay or boundary handling.
Weaker configurations fail earlier: Deepseek-Flash has 102/125 Unrecognizable instances, and the four lowest-scoring models produce no page in 12\%--19\% of instances.
Visual mismatch is the rarest category across configurations (2.3\% overall).
Under this gate order and these thresholds, replicas that meet the rule criteria thus usually also meet the visual criteria.

\subsubsection{Process-level Root-cause Attribution}
\label{sec:failure-submode}

\begin{figure*}[!t]
  \centering
  \includegraphics[width=\textwidth]{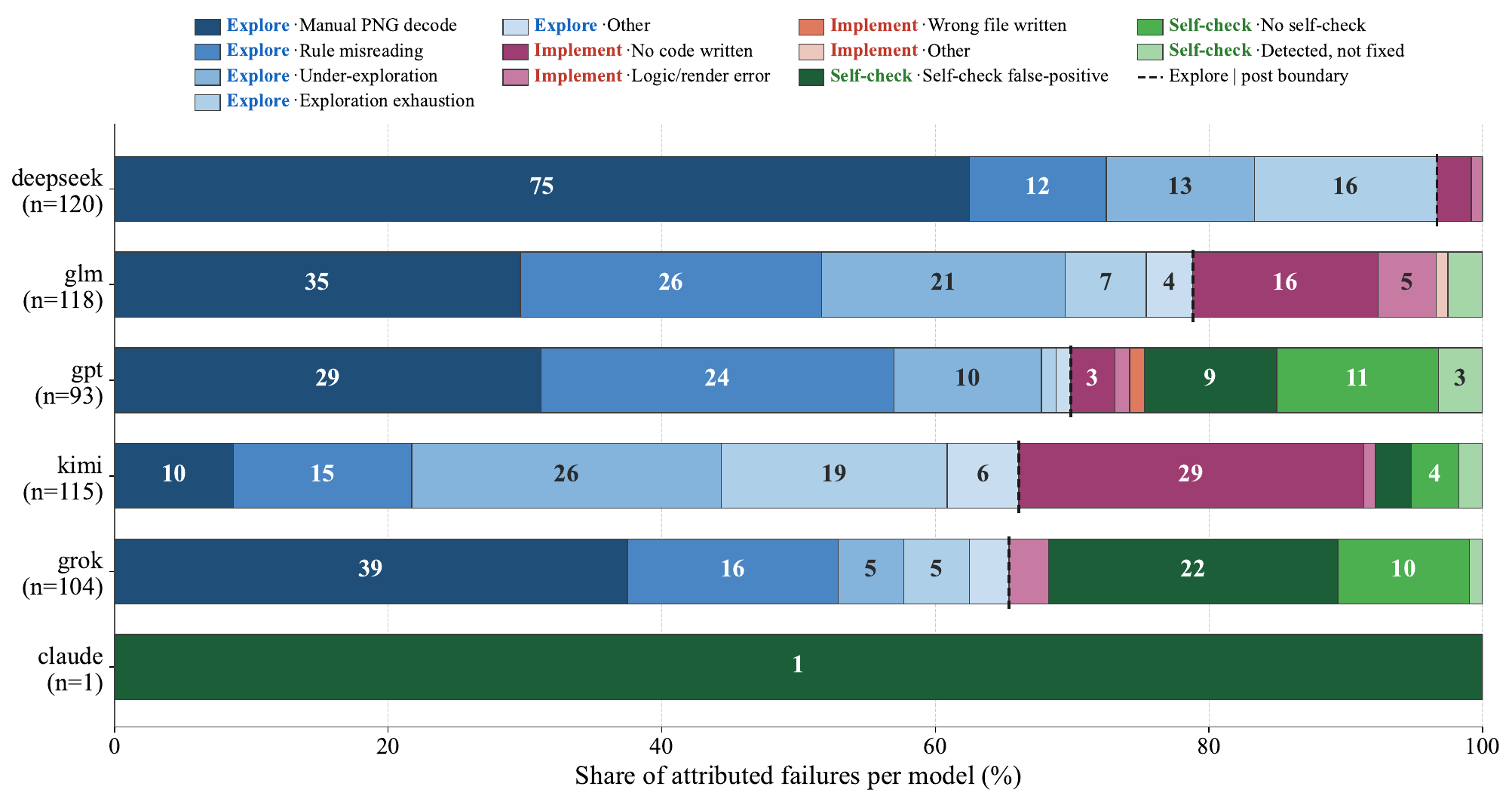}
  \caption{Process-level failure attribution of game replication. Blue, magenta, and green color families correspond to the exploration, implementation, and self-check phases, respectively; lighter and darker shades distinguish sub-modes within each phase. The dashed line on each bar marks the boundary between the exploration phase and subsequent phases.}
  \label{fig:failure-submode}
\end{figure*}

Trajectory comparison localizes attributed failures to the phase in which they originate.
The F5 reference pool covers 120 distinct game--difficulty pairs, with one successful trajectory per pair: 119 from Claude Opus 4.8 and one from GPT-5.5.
We compare these successful references with failed trajectories (F1--F4) through four steps:
\begin{itemize}
  \item \textbf{Sample definition}: We label all instances using the outcome criteria (Table~\ref{tab:failure-tiers}) and select F1--F4 instances for analysis. The attributed sample is the subset of these instances shown in Figure~\ref{fig:failure-submode}, which reports its size $n$ for each model.
  For each attributed failure, we use the F5 reference for the same game and difficulty level.
  \item \textbf{Phase segmentation}: We segment normalized trajectories using a state machine into three phases: exploration (observe the original game and infer rules), implementation (generate or edit replica code), and self-check (render and verify the replica).
  Implementation and self-check can recur across multiple iterations.
  \item \textbf{Root-cause phase attribution}: We ask an LLM judge to compare each failed trajectory with its matched successful trajectory and identify the phase where the failure originates.
  \item \textbf{Sub-mode refinement}: We ask the LLM to derive typical sub-modes within each phase and assign each failure to one sub-mode.
  The resulting model $\times$ phase $\times$ sub-mode distribution uses the categories in Table~\ref{tab:failure-submodes}, Appendix~\ref{app:failure-tables}.
\end{itemize}

\textbf{Most attributed failures originate in the exploration phase.}
As shown in Figure~\ref{fig:failure-submode}, failures among the five lower-scoring models predominantly precede implementation, occurring while agents observe and interpret the target game. The leading sub-mode, \emph{Manual PNG decode}, involves parsing screenshots pixel by pixel or byte by byte via scripts rather than viewing them directly, which discards crucial layout and gameplay cues. This sub-mode dominates the failures of Deepseek-Flash (36 of 60) and Grok v4.5 (23 of 52), and features prominently in GLM-5.3-Flash and GPT-5.5, followed by \emph{Rule misreading} and \emph{Under-exploration} as the other major exploration bottlenecks. Individual models also diverge in their broader failure profiles. Kimi K3 exhibits concentrated implementation-phase failures, failing to produce replica code in 13 of its 58 attributed cases. Meanwhile, GPT-5.5 advances to self-check more regularly before failing, whereas Grok v4.5 experiences frequent self-check errors driven largely by false-positive assessments of replica success. In sharp contrast, Claude Opus 4.8 incurs only a single attributed failure, namely a false-positive self-check. Within the attributed sample, weaker configurations thus fail primarily during pre-coding exploration before any replica implementation begins.

\section{Conclusion}
We introduced GameReplica, a complexity-controlled benchmark for black-box game replication that makes specification recovery an explicit requirement of executable reconstruction.
Its action--screenshot interface requires agents to infer gameplay rules and build runnable replicas across 25 games $\times$ 5 difficulty levels, totaling 125 instances.
Claude Opus 4.8 obtains an Overall score of 71.6, compared with 4.0--43.0 for the other models.
Across models, visual fidelity exceeds implementation consistency, which exceeds rule consistency.
Difficulty widens performance gaps: from L1 to L5, weaker models decline more than the strongest model.
The evaluated configurations vary more across models than across agent runtimes, and most attributed failures of weaker models arise in the exploration phase.
GameReplica enables future coding-agent research to measure whether improvements in active exploration and rule induction translate into more faithful executable replicas.

\section*{Acknowledgment}

We gratefully acknowledge Alibaba Group for its support of the experimental
work in this study, including the computational resources and experimental
infrastructure that facilitated the development and large-scale evaluation of
GameReplica. We also thank the colleagues who provided valuable discussions
and constructive feedback throughout this work.

\input{main.bbl}

\appendices

\section{Rule Categories and Complexity Metrics}
\label{app:rules-complexity}

Table~\ref{tab:rule-categories} defines the six categories of atomic rules that
make up the rule checklist $C$ used in
Section~\ref{sec:gamedesign}, and Table~\ref{tab:complexity-metrics} explains
the structural and solving complexity metrics used to characterize each
difficulty level.

\begin{table*}[t]\centering
\caption{Definitions of the six categories of atomic rules in the
rule checklist.}
\label{tab:rule-categories}
\fittable{%
\rowcolors{2}{sectionpurple!8}{white}
\begin{tabular*}{\textwidth}{@{\extracolsep{\fill}}>{\bfseries}m{0.28\textwidth} m{0.66\textwidth}}
\toprule
\rowcolor{sectionpurple!22}
\textbf{Rule category} & \textbf{Description} \\
\midrule
Visual & Static appearance: the shape, color, size, and relative position of
game elements and the overall UI layout, determining whether the replica's
screenshots look consistent with the original. \\
Legal action & The player operations allowed or forbidden in a given state, and
how each operation maps to a change in game state (e.g., the directions a piece
can move, which objects are interactive). \\
State & The definition and transition of the game's internal state: state
variables such as score, counters, positions, and level progress, together with
their update rules. \\
Boundary & Boundary and constraint conditions: board/map edges, collision
detection, out-of-bounds handling, and numeric lower/upper limits. \\
Terminal lock & Win/lose and end-of-game adjudication, and the locking behavior
after a terminal condition is reached (forbidding further operations, freezing
the screen, showing a settlement panel, etc.). \\
Level layout & The initial configuration of each difficulty instance: the
placement and count of initial elements and the map/board structure. \\
\bottomrule
\end{tabular*}%
}
\end{table*}

\begin{table*}[t]\centering
\caption{Structural and solving complexity metrics.}
\label{tab:complexity-metrics}
\fittable{%
\begin{tabular*}{\textwidth}{@{\extracolsep{\fill}}m{0.24\textwidth} >{\centering\arraybackslash}m{0.08\textwidth} m{0.52\textwidth}}
\toprule
\rowcolor{sectionpurple!22}
\textbf{Group} & \textbf{Metric} & \textbf{Description} \\
\midrule
\rowcolor{grpblue}  & $N_{\text{act}}$ & Number of candidate action
types available to the player; more actions imply a larger rule space. \\
\rowcolor{grpblue}  & $N_{\text{trans}}$ & Number of state-transition rules, i.e., the entries
specifying ``how actions change the state.'' \\
\rowcolor{grpblue} \multirow{-5}{0.23\textwidth}{\textbf{Rule complexity (structural)}} & $N_{\text{goal}}$ & Number of goal/win-lose/termination rules, characterizing
the complexity of terminal adjudication. \\
\cmidrule(l{2pt}r{2pt}){1-3}
\rowcolor{grpgreen}  & $N_{\text{unit}}$ & Number of state units (e.g.,
cells, slots) that make up the game state, determining the size of the state
space. \\
\rowcolor{grpgreen} \multirow{-3}{0.23\textwidth}{\textbf{State complexity (structural)}} & $N_{\text{obj}}$ & Number of dynamic object types in the state; more types
imply richer state combinations. \\
\cmidrule(l{2pt}r{2pt}){1-3}
\rowcolor{grpamber}  & $N_{\text{step}}$ & Length of the shortest action sequence
from the initial state to the goal state, i.e., the minimal-solution depth. \\
\rowcolor{grpamber} \multirow{-4}{0.23\textwidth}[-6.6pt]{\textbf{Solving complexity}} & $\bar{B}$ & Average number of legal actions over reachable states, i.e., the
mean branching factor, reflecting the search width. \\
\bottomrule
\end{tabular*}%
}
\end{table*}

\section{Failure Analysis Tables}
\label{app:failure-tables}

Table~\ref{tab:failure-tiers} defines the five-tier outcome taxonomy used for
the outcome-level failure decomposition in Section~\ref{sec:failure-decomp}, and
Table~\ref{tab:failure-submodes} lists the process-level failure sub-modes used
in Section~\ref{sec:failure-submode}.

\begin{table*}[t]
\centering
\caption{Five-tier outcome taxonomy and score ranges for replication results. Each instance is assigned along an ordered funnel to its first unmet gate; instances that clear all four gates are Faithful. The classification is mutually exclusive and exhaustive, so each configuration's tier shares sum to 100\%.}
\label{tab:failure-tiers}
\setlength{\tabcolsep}{4pt}
\small
\fittable{%
\rowcolors{2}{sectionpurple!8}{white}
\begin{tabular*}{\textwidth}{@{\extracolsep{\fill}}@{}>{\bfseries}m{0.08\textwidth}>{\bfseries}m{0.17\textwidth}>{\centering\arraybackslash}m{0.07\textwidth}>{\centering\arraybackslash}m{0.09\textwidth}>{\centering\arraybackslash}m{0.15\textwidth}m{0.28\textwidth}@{}}
\toprule
\rowcolor{sectionpurple!22}
\textbf{Tier} & \textbf{Failure mode} & \textbf{HTML} & \textbf{Visual} & \textbf{$S^{\text{mech}}$} & \textbf{Meaning} \\
\midrule
F1 & No output       & None & ---        & ---          & No page artifact produced; failed at the front gate. \\
F2 & Unrecognizable  & Yes  & $[0,20)$   & n/a          & Visual fidelity too low: screenshot is blank / error / skeleton; the game is unrecognizable. \\
F3 & Rule violation  & Yes  & $\geq 20$  & $[0,50)$     & Game is recognizable but the mechanism score falls short (gameplay / win-loss / boundary errors). \\
F4 & Visual mismatch & Yes  & $[20,60)$  & $[50,100]$   & The mechanism score passes, but visual fidelity remains below the threshold. \\
F5 & Faithful        & Yes  & $[60,100]$ & $[50,100]$   & Visual fidelity and the mechanism score both pass. \\
\bottomrule
\end{tabular*}%
}
\end{table*}

\begin{table*}[t]
\centering
\caption{Failure sub-modes organized by replication phase. Each sub-mode is inductively derived from LLM-judge attribution rationales across all failed instances.}
\label{tab:failure-submodes}
\setlength{\tabcolsep}{3pt}
\small
\fittable{%
\begin{tabular*}{\textwidth}{@{\extracolsep{\fill}}@{}m{0.15\textwidth}>{\bfseries}m{0.24\textwidth}m{0.53\textwidth}@{}}
\toprule
\rowcolor{sectionpurple!22}
\textbf{Phase} & \textbf{Sub-mode} & \textbf{Description} \\
\midrule
\rowcolor{grpblue}   & Manual PNG decode      & Instead of viewing the game directly, the agent writes scripts to decode screenshots byte-by-byte / pixel-by-pixel, getting stuck in low-level image parsing and losing layout and gameplay information. \\
\rowcolor{grpblue}   & Rule misreading        & The agent misunderstands the type / mechanics / rules of the original game, mistaking the game or inducing a biased rule set from the outset. \\
\rowcolor{grpblue}   & Under-exploration      & The agent observes or interacts too little and enters implementation prematurely, without grasping the game's rules and appearance. \\
\rowcolor{grpblue}   & Exploration exhaustion & The agent burns a large number of steps / budget during exploration (dead loops, ineffective attempts), leaving no resources to advance to implementation. \\
\rowcolor{grpblue} \multirow{-12}{*}{\textbf{Exploration}}
  & Other & Exploration failures that do not fall into the sub-modes above. \\
\midrule
\rowcolor{grpgreen}   & No code written        & The agent never writes any replica page code, often because it is stuck in exploration and never enters implementation. \\
\rowcolor{grpgreen}   & Logic/render error     & Code is written but its logic, styling, layout, or interaction is wrong, so the page renders incorrectly or incompletely. \\
\rowcolor{grpgreen}   & Wrong file written     & Content is written to the wrong file (e.g., only editing the game\_spec document rather than writing the replica HTML). \\
\rowcolor{grpgreen} \multirow{-7}{*}{\textbf{Implementation}}
  & Other & Implementation failures that do not fall into the sub-modes above. \\
\midrule
\rowcolor{grpamber}   & Self-check false-positive & During self-check the agent wrongly judges success, submitting without noticing that the page is malformed / errors out / deviates from the original. \\
\rowcolor{grpamber}   & No self-check          & The agent never renders and verifies the replica page, skipping self-check and finishing directly. \\
\rowcolor{grpamber} \multirow{-6}{*}{\textbf{Self-check}}
  & Detected, not fixed    & The agent identifies a problem but never attempts to fix it (never triggers the re-edit loop). \\
\bottomrule
\end{tabular*}%
}
\end{table*}

\section{Game Categories}
\label{app:game-categories}

Table~\ref{tab:game-category-mechanisms} summarizes the five game categories,
their core gameplay rules, and representative games.

\begin{table*}[!t]
\centering
\caption{Puzzle and game categories with their core mechanisms and representative games.}
\label{tab:game-category-mechanisms}
\begingroup
\small
\setlength{\tabcolsep}{8pt}
\renewcommand{\arraystretch}{1.35}
\setlength{\extrarowheight}{4pt}
\rowcolors{2}{black!3}{white}
\begin{tabularx}{0.96\textwidth}{
  >{\hsize=0.78\hsize\linewidth=\hsize\raggedright\arraybackslash\bfseries}X
  >{\hsize=1.17\hsize\linewidth=\hsize\raggedright\arraybackslash}X
  >{\hsize=1.05\hsize\linewidth=\hsize\raggedright\arraybackslash}X}
\toprule[0.9pt]
\rowcolor{black!9}
\textbf{Category} & \textbf{Core Mechanism} & \textbf{Representative Games} \\[4pt]
\midrule[0.45pt]
Spatial Rearrangement and Pattern Transformation & Move objects or toggle cell states to achieve a target configuration. & Sokoban, Rush Hour, Nonogram, Lights Out, Klotski \\
Path Connection and Network Construction & Construct paths, bridges, or loops to satisfy connectivity constraints. & Color Connect, Hashiwokakero, Loops, NetWalk, Slitherlink \\
Turn-Based Board Games & Alternate turns to compete through capture, flipping, blocking, or alignment. & Reversi, Nine Men's Morris, Checkers, Connect Four, Gomoku \\
Sequential Planning and Resource Transfer & Plan object transfers under ordering, capacity, or safety constraints. & Hanoi, Water Jug, Ball Sort, Leapfrog, River Crossing \\
Grid Constraints and Logical Filling & Fill grids to satisfy regional, inequality, arithmetic, or counting constraints. & Suguru, Futoshiki, Skyscrapers, KenKen, Takuzu \\
\bottomrule[0.9pt]
\end{tabularx}
\endgroup
\end{table*}

\section{Per-Configuration Difficulty Decay}
\label{app:difficulty-decay}

Tables~\ref{tab:decay-ClaudeOpus}--\ref{tab:decay-OpenHands} report per-level scores and resource use for each configuration.
Scores are mean $\pm$ sample standard deviation over the 25 instances at each level ($\times$100), covering rule consistency, implementation consistency, visual fidelity, and the Overall score.
The \textit{All} row aggregates all 125 instances.
The model comparison fixes the agent runtime to Claude Code; the agent runtime comparison fixes the model to Claude Opus 4.8.
The Claude Code and Claude Opus 4.8 tables therefore report the same shared configuration.

\subsection*{Model Comparison (agent runtime fixed to Claude Code)}

Tables~\ref{tab:decay-ClaudeOpus}--\ref{tab:decay-Deepseek} report the six models with the agent runtime fixed to Claude Code.

\begin{table*}[t]
\centering
\caption{Difficulty decay for \textbf{Claude Opus 4.8}. Overall score = equal-weight mean of rule consistency, implementation consistency, and visual fidelity ($\times$100). Scores are percentages, reported as mean $\pm$ sample standard deviation across the 25 instances per level; \textit{All} pools the 125 instances. Each instance score averages three independent runs. Bold numbers mark pooled results. Resource statistics use successfully produced instances.}
\label{tab:decay-ClaudeOpus}
\setlength{\tabcolsep}{5pt}
\fittable{%
\begin{tabular*}{\textwidth}{@{\extracolsep{\fill}}lcccccc}
\toprule
\textbf{Level} & \textbf{Rule} & \textbf{Impl} & \textbf{Visual} & \textbf{Overall} & \textbf{Tokens (M)} & \textbf{Time (s)} \\
\midrule
L1 & \cellcolor[rgb]{0.571,0.792,0.645} $54.8\pm14.7$& \cellcolor[rgb]{0.529,0.619,0.861} $68.2\pm8.3$& \cellcolor[rgb]{0.744,0.583,0.771} $93.2\pm4.4$& \cellcolor[rgb]{0.669,0.669,0.669} $72.1\pm8.0$& $3.27\pm2.58$ & $1122\pm916$ \\
L2 & \cellcolor[rgb]{0.3,0.66,0.42} $57.0\pm10.4$& \cellcolor[rgb]{0.32,0.45,0.8} $71.1\pm6.7$& \cellcolor[rgb]{0.677,0.474,0.711} $93.9\pm3.0$& \cellcolor[rgb]{0.52,0.52,0.52} $74.0\pm4.3$& $3.73\pm1.79$ & $1290\pm747$ \\
L3 & \cellcolor[rgb]{0.583,0.798,0.655} $54.7\pm9.0$& \cellcolor[rgb]{0.414,0.526,0.828} $69.8\pm9.3$& \cellcolor[rgb]{0.62,0.38,0.66} $94.5\pm4.7$& \cellcolor[rgb]{0.598,0.598,0.598} $73.0\pm5.2$& $4.96\pm3.93$ & $1625\pm1279$ \\
L4 & \cellcolor[rgb]{0.842,0.923,0.869} $52.6\pm11.9$& \cellcolor[rgb]{0.753,0.8,0.927} $65.1\pm11.1$& \cellcolor[rgb]{0.763,0.614,0.788} $93.0\pm6.6$& \cellcolor[rgb]{0.817,0.817,0.817} $70.2\pm8.1$& $5.55\pm2.97$ & $1860\pm926$ \\
L5 & \cellcolor[rgb]{0.916,0.959,0.93} $52.0\pm13.4$& \cellcolor[rgb]{0.918,0.934,0.976} $62.8\pm13.9$& \cellcolor[rgb]{0.954,0.926,0.959} $91.0\pm7.4$& \cellcolor[rgb]{0.942,0.942,0.942} $68.6\pm9.6$& $7.72\pm3.04$ & $2576\pm1300$ \\
\midrule
\textit{All} & $\mathbf{54.2\pm12.0}$ & $\mathbf{67.4\pm10.5}$ & $\mathbf{93.1\pm5.5}$ & $\mathbf{71.6\pm7.4}$ & $\mathbf{5.05\pm3.30}$ & $\mathbf{1694\pm1159}$ \\
\bottomrule
\end{tabular*}%
}
\end{table*}

\begin{table*}[t]
\centering
\caption{Difficulty decay for \textbf{GPT-5.5}. Scores are percentages, reported as mean $\pm$ sample standard deviation across the 25 instances per level; \textit{All} pools the 125 instances. Each instance score averages three independent runs. Bold numbers mark pooled results. Resource statistics use successfully produced instances.}
\label{tab:decay-GPT}
\setlength{\tabcolsep}{5pt}
\fittable{%
\begin{tabular*}{\textwidth}{@{\extracolsep{\fill}}lcccccc}
\toprule
\textbf{Level} & \textbf{Rule} & \textbf{Impl} & \textbf{Visual} & \textbf{Overall} & \textbf{Tokens (M)} & \textbf{Time (s)} \\
\midrule
L1 & \cellcolor[rgb]{0.336,0.677,0.45} $31.9\pm15.7$& \cellcolor[rgb]{0.42,0.531,0.829} $45.8\pm15.2$& \cellcolor[rgb]{0.765,0.617,0.79} $62.2\pm22.1$& \cellcolor[rgb]{0.59,0.59,0.59} $46.6\pm11.6$& $2.04\pm1.92$ & $1267\pm261$ \\
L2 & \cellcolor[rgb]{0.3,0.66,0.42} $32.8\pm13.5$& \cellcolor[rgb]{0.32,0.45,0.8} $49.3\pm16.8$& \cellcolor[rgb]{0.717,0.538,0.747} $63.4\pm25.4$& \cellcolor[rgb]{0.52,0.52,0.52} $48.5\pm14.9$& $2.24\pm2.35$ & $1370\pm407$ \\
L3 & \cellcolor[rgb]{0.622,0.816,0.687} $24.7\pm13.5$& \cellcolor[rgb]{0.609,0.684,0.885} $39.2\pm15.6$& \cellcolor[rgb]{0.954,0.926,0.959} $57.5\pm18.0$& \cellcolor[rgb]{0.814,0.814,0.814} $40.5\pm13.3$& $2.09\pm1.70$ & $1184\pm230$ \\
L4 & \cellcolor[rgb]{0.916,0.959,0.93} $17.3\pm14.2$& \cellcolor[rgb]{0.918,0.934,0.976} $28.4\pm18.3$& \cellcolor[rgb]{0.644,0.419,0.682} $65.2\pm21.1$& \cellcolor[rgb]{0.942,0.942,0.942} $37.0\pm15.6$& $2.46\pm1.79$ & $1444\pm456$ \\
L5 & \cellcolor[rgb]{0.554,0.784,0.631} $26.4\pm19.6$& \cellcolor[rgb]{0.747,0.795,0.925} $34.4\pm23.3$& \cellcolor[rgb]{0.62,0.38,0.66} $65.8\pm19.9$& \cellcolor[rgb]{0.751,0.751,0.751} $42.2\pm19.6$& $2.81\pm2.17$ & $1510\pm306$ \\
\midrule
\textit{All} & $\mathbf{26.6\pm16.2}$ & $\mathbf{39.4\pm19.3}$ & $\mathbf{62.8\pm21.3}$ & $\mathbf{43.0\pm15.6}$ & $\mathbf{2.33\pm1.99}$ & $\mathbf{1355\pm358}$ \\
\bottomrule
\end{tabular*}%
}
\end{table*}

\begin{table*}[t]
\centering
\caption{Difficulty decay for \textbf{Grok v4.5}. Scores are percentages, reported as mean $\pm$ sample standard deviation across the 25 instances per level; \textit{All} pools the 125 instances. Each instance score averages three independent runs. Bold numbers mark pooled results. Resource statistics use successfully produced instances.}
\label{tab:decay-Grok}
\setlength{\tabcolsep}{5pt}
\fittable{%
\begin{tabular*}{\textwidth}{@{\extracolsep{\fill}}lcccccc}
\toprule
\textbf{Level} & \textbf{Rule} & \textbf{Impl} & \textbf{Visual} & \textbf{Overall} & \textbf{Tokens (M)} & \textbf{Time (s)} \\
\midrule
L1 & \cellcolor[rgb]{0.3,0.66,0.42} $21.4\pm15.4$& \cellcolor[rgb]{0.32,0.45,0.8} $30.6\pm18.9$& \cellcolor[rgb]{0.62,0.38,0.66} $55.6\pm20.7$& \cellcolor[rgb]{0.52,0.52,0.52} $35.9\pm15.3$& $10.73\pm5.80$ & $1582\pm522$ \\
L2 & \cellcolor[rgb]{0.35,0.684,0.461} $20.2\pm19.4$& \cellcolor[rgb]{0.385,0.503,0.819} $28.1\pm23.2$& \cellcolor[rgb]{0.79,0.658,0.812} $43.8\pm28.3$& \cellcolor[rgb]{0.628,0.628,0.628} $30.7\pm21.0$& $10.60\pm6.15$ & $1608\pm713$ \\
L3 & \cellcolor[rgb]{0.478,0.746,0.567} $17.1\pm15.6$& \cellcolor[rgb]{0.453,0.557,0.839} $25.5\pm22.1$& \cellcolor[rgb]{0.804,0.681,0.825} $42.8\pm28.8$& \cellcolor[rgb]{0.673,0.673,0.673} $28.5\pm19.9$& $12.30\pm7.72$ & $1649\pm617$ \\
L4 & \cellcolor[rgb]{0.573,0.793,0.646} $14.8\pm17.8$& \cellcolor[rgb]{0.562,0.646,0.871} $21.3\pm21.8$& \cellcolor[rgb]{0.694,0.5,0.726} $50.5\pm22.2$& \cellcolor[rgb]{0.667,0.667,0.667} $28.8\pm17.6$& $13.50\pm9.47$ & $1904\pm793$ \\
L5 & \cellcolor[rgb]{0.916,0.959,0.93} $6.5\pm9.2$& \cellcolor[rgb]{0.918,0.934,0.976} $7.6\pm10.4$& \cellcolor[rgb]{0.954,0.926,0.959} $32.4\pm25.9$& \cellcolor[rgb]{0.942,0.942,0.942} $15.5\pm13.8$& $13.81\pm8.89$ & $1874\pm731$ \\
\midrule
\textit{All} & $\mathbf{16.0\pm16.5}$ & $\mathbf{22.6\pm21.1}$ & $\mathbf{45.0\pm26.2}$ & $\mathbf{27.9\pm18.7}$ & $\mathbf{12.19\pm7.74}$ & $\mathbf{1723\pm685}$ \\
\bottomrule
\end{tabular*}%
}
\end{table*}

\begin{table*}[t]
\centering
\caption{Difficulty decay for \textbf{GLM-5.3-Flash}. Scores are percentages, reported as mean $\pm$ sample standard deviation across the 25 instances per level; \textit{All} pools the 125 instances. Each instance score averages three independent runs. Bold numbers mark pooled results. Resource statistics use successfully produced instances.}
\label{tab:decay-GLM}
\setlength{\tabcolsep}{5pt}
\fittable{%
\begin{tabular*}{\textwidth}{@{\extracolsep{\fill}}lcccccc}
\toprule
\textbf{Level} & \textbf{Rule} & \textbf{Impl} & \textbf{Visual} & \textbf{Overall} & \textbf{Tokens (M)} & \textbf{Time (s)} \\
\midrule
L1 & \cellcolor[rgb]{0.3,0.66,0.42} $13.6\pm17.9$& \cellcolor[rgb]{0.32,0.45,0.8} $20.7\pm21.9$& \cellcolor[rgb]{0.62,0.38,0.66} $41.3\pm25.2$& \cellcolor[rgb]{0.52,0.52,0.52} $25.2\pm18.1$& $16.66\pm11.31$ & $3934\pm1174$ \\
L2 & \cellcolor[rgb]{0.318,0.669,0.435} $13.3\pm15.5$& \cellcolor[rgb]{0.365,0.487,0.813} $19.7\pm17.4$& \cellcolor[rgb]{0.651,0.431,0.688} $39.2\pm26.4$& \cellcolor[rgb]{0.556,0.556,0.556} $24.0\pm16.2$& $16.05\pm12.30$ & $3779\pm1349$ \\
L3 & \cellcolor[rgb]{0.729,0.868,0.775} $6.5\pm7.2$& \cellcolor[rgb]{0.764,0.809,0.931} $10.9\pm12.1$& \cellcolor[rgb]{0.771,0.626,0.795} $31.2\pm22.9$& \cellcolor[rgb]{0.788,0.788,0.788} $16.2\pm11.8$& $16.14\pm10.16$ & $3835\pm1471$ \\
L4 & \cellcolor[rgb]{0.916,0.959,0.93} $3.4\pm5.7$& \cellcolor[rgb]{0.855,0.883,0.957} $8.9\pm12.4$& \cellcolor[rgb]{0.848,0.753,0.864} $26.0\pm23.7$& \cellcolor[rgb]{0.889,0.889,0.889} $12.8\pm11.6$& $18.73\pm10.26$ & $4213\pm1481$ \\
L5 & \cellcolor[rgb]{0.729,0.868,0.775} $6.5\pm13.0$& \cellcolor[rgb]{0.918,0.934,0.976} $7.5\pm15.2$& \cellcolor[rgb]{0.954,0.926,0.959} $18.9\pm19.1$& \cellcolor[rgb]{0.942,0.942,0.942} $11.0\pm13.6$& $18.04\pm9.44$ & $3925\pm1518$ \\
\midrule
\textit{All} & $\mathbf{8.6\pm13.2}$ & $\mathbf{13.6\pm16.9}$ & $\mathbf{31.3\pm24.6}$ & $\mathbf{17.8\pm15.4}$ & $\mathbf{17.12\pm10.62}$ & $\mathbf{3937\pm1390}$ \\
\bottomrule
\end{tabular*}%
}
\end{table*}

\begin{table*}[t]
\centering
\caption{Difficulty decay for \textbf{Kimi K3}. Scores are percentages, reported as mean $\pm$ sample standard deviation across the 25 instances per level; \textit{All} pools the 125 instances. Each instance score averages three independent runs. Bold numbers mark pooled results. Resource statistics use successfully produced instances.}
\label{tab:decay-Kimi}
\setlength{\tabcolsep}{5pt}
\fittable{%
\begin{tabular*}{\textwidth}{@{\extracolsep{\fill}}lcccccc}
\toprule
\textbf{Level} & \textbf{Rule} & \textbf{Impl} & \textbf{Visual} & \textbf{Overall} & \textbf{Tokens (M)} & \textbf{Time (s)} \\
\midrule
L1 & \cellcolor[rgb]{0.3,0.66,0.42} $6.4\pm10.7$& \cellcolor[rgb]{0.32,0.45,0.8} $11.9\pm21.9$& \cellcolor[rgb]{0.713,0.531,0.743} $28.0\pm24.6$& \cellcolor[rgb]{0.52,0.52,0.52} $15.4\pm17.2$& $23.91\pm11.28$ & $2291\pm785$ \\
L2 & \cellcolor[rgb]{0.496,0.755,0.582} $5.0\pm8.4$& \cellcolor[rgb]{0.514,0.607,0.857} $9.6\pm13.4$& \cellcolor[rgb]{0.62,0.38,0.66} $30.3\pm27.6$& \cellcolor[rgb]{0.553,0.553,0.553} $15.0\pm14.7$& $8.03\pm9.86$   & $1022\pm965$ \\
L3 & \cellcolor[rgb]{0.916,0.959,0.93} $2.0\pm3.0$& \cellcolor[rgb]{0.859,0.886,0.959} $5.5\pm7.3$& \cellcolor[rgb]{0.656,0.439,0.692} $29.4\pm21.4$& \cellcolor[rgb]{0.777,0.777,0.777} $12.3\pm8.9$& $15.39\pm10.22$ & $1812\pm964$ \\
L4 & \cellcolor[rgb]{0.622,0.816,0.687} $4.1\pm8.9$& \cellcolor[rgb]{0.775,0.818,0.934} $6.5\pm13.3$& \cellcolor[rgb]{0.954,0.926,0.959} $22.0\pm25.9$& \cellcolor[rgb]{0.893,0.893,0.893} $10.9\pm14.4$& $22.66\pm15.41$ & $2512\pm1210$ \\
L5 & \cellcolor[rgb]{0.664,0.837,0.722} $3.8\pm6.0$& \cellcolor[rgb]{0.918,0.934,0.976} $4.8\pm8.3$& \cellcolor[rgb]{0.942,0.906,0.948} $22.3\pm27.8$& \cellcolor[rgb]{0.942,0.942,0.942} $10.3\pm13.1$& $32.17\pm21.03$ & $2261\pm915$ \\
\midrule
\textit{All} & $\mathbf{4.2\pm7.9}$ & $\mathbf{7.6\pm13.9}$ & $\mathbf{26.4\pm25.4}$ & $\mathbf{12.8\pm13.9}$ & $\mathbf{20.98\pm15.71}$ & $\mathbf{1982\pm1010}$ \\
\bottomrule
\end{tabular*}%
}
\end{table*}

\begin{table*}[t]
\centering
\caption{Difficulty decay for \textbf{Deepseek-Flash}. Scores are percentages, reported as mean $\pm$ sample standard deviation across the 25 instances per level; \textit{All} pools the 125 instances. Each instance score averages three independent runs. Bold numbers mark pooled results. Resource statistics use successfully produced instances.}
\label{tab:decay-Deepseek}
\setlength{\tabcolsep}{5pt}
\fittable{%
\begin{tabular*}{\textwidth}{@{\extracolsep{\fill}}lcccccc}
\toprule
\textbf{Level} & \textbf{Rule} & \textbf{Impl} & \textbf{Visual} & \textbf{Overall} & \textbf{Tokens (M)} & \textbf{Time (s)} \\
\midrule
L1 & \cellcolor[rgb]{0.3,0.66,0.42} $2.5\pm7.4$& \cellcolor[rgb]{0.32,0.45,0.8} $6.4\pm9.8$& \cellcolor[rgb]{0.758,0.605,0.783} $7.2\pm12.7$& \cellcolor[rgb]{0.52,0.52,0.52} $5.4\pm9.3$& $14.01\pm6.34$ & $2642\pm778$ \\
L2 & \cellcolor[rgb]{0.755,0.881,0.797} $0.8\pm1.9$& \cellcolor[rgb]{0.554,0.639,0.869} $4.6\pm7.7$& \cellcolor[rgb]{0.753,0.597,0.779} $7.3\pm11.8$& \cellcolor[rgb]{0.669,0.669,0.669} $4.2\pm5.1$& $17.60\pm4.56$ & $3083\pm400$ \\
L3 & \cellcolor[rgb]{0.648,0.829,0.708} $1.2\pm2.3$& \cellcolor[rgb]{0.762,0.808,0.93} $3.0\pm4.3$& \cellcolor[rgb]{0.709,0.524,0.739} $8.2\pm16.8$& \cellcolor[rgb]{0.682,0.682,0.682} $4.1\pm6.6$& $15.42\pm4.41$ & $2855\pm670$ \\
L4 & \cellcolor[rgb]{0.916,0.959,0.93} $0.2\pm0.6$& \cellcolor[rgb]{0.918,0.934,0.976} $1.8\pm3.4$& \cellcolor[rgb]{0.62,0.38,0.66} $10.0\pm14.1$& \cellcolor[rgb]{0.694,0.694,0.694} $4.0\pm5.2$& $18.87\pm8.47$ & $3137\pm332$ \\
L5 & \cellcolor[rgb]{0.782,0.894,0.819} $0.7\pm2.9$& \cellcolor[rgb]{0.892,0.913,0.968} $2.0\pm4.4$& \cellcolor[rgb]{0.954,0.926,0.959} $3.2\pm6.7$& \cellcolor[rgb]{0.942,0.942,0.942} $2.0\pm4.4$& $17.37\pm5.09$ & $2944\pm689$ \\
\midrule
\textit{All} & $\mathbf{1.1\pm3.8}$ & $\mathbf{3.6\pm6.5}$ & $\mathbf{7.2\pm12.8}$ & $\mathbf{4.0\pm6.4}$ & $\mathbf{16.42\pm5.94}$ & $\mathbf{2912\pm614}$ \\
\bottomrule
\end{tabular*}%
}
\end{table*}

\subsection*{Agent Runtime Comparison (model fixed to Claude Opus 4.8)}

Tables~\ref{tab:decay-ClaudeCode}--\ref{tab:decay-OpenHands} report the five agent runtimes with the model fixed to Claude Opus 4.8.

\begin{table*}[t]
\centering
\caption{Difficulty decay for \textbf{Claude Code} agent runtime. Values are identical to Table~\ref{tab:decay-ClaudeOpus} because both tables report the same configuration. Scores are percentages, reported as mean $\pm$ sample standard deviation across the 25 instances per level; \textit{All} pools the 125 instances. Each instance score averages three independent runs. Bold numbers mark pooled results. Resource statistics use successfully produced instances.}
\label{tab:decay-ClaudeCode}
\setlength{\tabcolsep}{5pt}
\fittable{%
\begin{tabular*}{\textwidth}{@{\extracolsep{\fill}}lcccccc}
\toprule
\textbf{Level} & \textbf{Rule} & \textbf{Impl} & \textbf{Visual} & \textbf{Overall} & \textbf{Tokens (M)} & \textbf{Time (s)} \\
\midrule
L1 & \cellcolor[rgb]{0.571,0.792,0.645} $54.8\pm14.7$& \cellcolor[rgb]{0.529,0.619,0.861} $68.2\pm8.3$& \cellcolor[rgb]{0.744,0.583,0.771} $93.2\pm4.4$& \cellcolor[rgb]{0.669,0.669,0.669} $72.1\pm8.0$& $3.27\pm2.58$ & $1122\pm916$ \\
L2 & \cellcolor[rgb]{0.3,0.66,0.42} $57.0\pm10.4$& \cellcolor[rgb]{0.32,0.45,0.8} $71.1\pm6.7$& \cellcolor[rgb]{0.677,0.474,0.711} $93.9\pm3.0$& \cellcolor[rgb]{0.52,0.52,0.52} $74.0\pm4.3$& $3.73\pm1.79$ & $1290\pm747$ \\
L3 & \cellcolor[rgb]{0.583,0.798,0.655} $54.7\pm9.0$& \cellcolor[rgb]{0.414,0.526,0.828} $69.8\pm9.3$& \cellcolor[rgb]{0.62,0.38,0.66} $94.5\pm4.7$& \cellcolor[rgb]{0.598,0.598,0.598} $73.0\pm5.2$& $4.96\pm3.93$ & $1625\pm1279$ \\
L4 & \cellcolor[rgb]{0.842,0.923,0.869} $52.6\pm11.9$& \cellcolor[rgb]{0.753,0.8,0.927} $65.1\pm11.1$& \cellcolor[rgb]{0.763,0.614,0.788} $93.0\pm6.6$& \cellcolor[rgb]{0.817,0.817,0.817} $70.2\pm8.1$& $5.55\pm2.97$ & $1860\pm926$ \\
L5 & \cellcolor[rgb]{0.916,0.959,0.93} $52.0\pm13.4$& \cellcolor[rgb]{0.918,0.934,0.976} $62.8\pm13.9$& \cellcolor[rgb]{0.954,0.926,0.959} $91.0\pm7.4$& \cellcolor[rgb]{0.942,0.942,0.942} $68.6\pm9.6$& $7.72\pm3.04$ & $2576\pm1300$ \\
\midrule
\textit{All} & $\mathbf{54.2\pm12.0}$ & $\mathbf{67.4\pm10.5}$ & $\mathbf{93.1\pm5.5}$ & $\mathbf{71.6\pm7.4}$ & $\mathbf{5.05\pm3.30}$ & $\mathbf{1694\pm1159}$ \\
\bottomrule
\end{tabular*}%
}
\end{table*}

\begin{table*}[t]
\centering
\caption{Difficulty decay for \textbf{Codex} agent runtime. Scores are percentages, reported as mean $\pm$ sample standard deviation across the 25 instances per level; \textit{All} pools the 125 instances. Each instance score averages three independent runs. Bold numbers mark pooled results. Resource statistics use successfully produced instances.}
\label{tab:decay-Codex}
\setlength{\tabcolsep}{5pt}
\fittable{%
\begin{tabular*}{\textwidth}{@{\extracolsep{\fill}}lcccccc}
\toprule
\textbf{Level} & \textbf{Rule} & \textbf{Impl} & \textbf{Visual} & \textbf{Overall} & \textbf{Tokens (M)} & \textbf{Time (s)} \\
\midrule
L1 & \cellcolor[rgb]{0.3,0.66,0.42} $52.3\pm11.6$& \cellcolor[rgb]{0.346,0.471,0.808} $66.2\pm9.1$& \cellcolor[rgb]{0.647,0.424,0.684} $92.2\pm3.9$& \cellcolor[rgb]{0.52,0.52,0.52} $70.2\pm6.6$& $2.54\pm1.30$ & $1596\pm707$ \\
L2 & \cellcolor[rgb]{0.3,0.66,0.42} $52.3\pm11.2$& \cellcolor[rgb]{0.424,0.534,0.831} $65.3\pm8.0$& \cellcolor[rgb]{0.62,0.38,0.66} $92.4\pm4.6$& \cellcolor[rgb]{0.54,0.54,0.54} $70.0\pm6.1$& $3.18\pm2.12$ & $1904\pm1083$ \\
L3 & \cellcolor[rgb]{0.317,0.668,0.434} $52.1\pm10.9$& \cellcolor[rgb]{0.32,0.45,0.8} $66.5\pm11.3$& \cellcolor[rgb]{0.954,0.926,0.959} $89.9\pm11.2$& \cellcolor[rgb]{0.589,0.589,0.589} $69.5\pm7.7$& $4.66\pm2.28$ & $2721\pm1342$ \\
L4 & \cellcolor[rgb]{0.916,0.959,0.93} $45.0\pm14.3$& \cellcolor[rgb]{0.65,0.717,0.897} $62.7\pm13.1$& \cellcolor[rgb]{0.941,0.904,0.947} $90.0\pm6.9$& \cellcolor[rgb]{0.942,0.942,0.942} $65.9\pm9.0$& $4.94\pm1.90$ & $2743\pm873$ \\
L5 & \cellcolor[rgb]{0.756,0.881,0.798} $46.9\pm14.0$& \cellcolor[rgb]{0.918,0.934,0.976} $59.6\pm13.3$& \cellcolor[rgb]{0.754,0.598,0.78} $91.4\pm8.0$& \cellcolor[rgb]{0.942,0.942,0.942} $65.9\pm9.0$& $5.56\pm1.85$ & $3131\pm1276$ \\
\midrule
\textit{All} & $\mathbf{49.7\pm12.7}$ & $\mathbf{64.0\pm11.3}$ & $\mathbf{91.2\pm7.4}$ & $\mathbf{68.3\pm7.9}$ & $\mathbf{4.18\pm2.20}$ & $\mathbf{2419\pm1211}$ \\
\bottomrule
\end{tabular*}%
}
\end{table*}

\begin{table*}[t]
\centering
\caption{Difficulty decay for \textbf{Hermes} agent runtime. Scores are percentages, reported as mean $\pm$ sample standard deviation across the 25 instances per level; \textit{All} pools the 125 instances. Each instance score averages three independent runs. Bold numbers mark pooled results. Resource statistics use successfully produced instances.}
\label{tab:decay-Hermes}
\setlength{\tabcolsep}{5pt}
\fittable{%
\begin{tabular*}{\textwidth}{@{\extracolsep{\fill}}lcccccc}
\toprule
\textbf{Level} & \textbf{Rule} & \textbf{Impl} & \textbf{Visual} & \textbf{Overall} & \textbf{Tokens (M)} & \textbf{Time (s)} \\
\midrule
L1 & \cellcolor[rgb]{0.329,0.674,0.444} $53.2\pm12.4$& \cellcolor[rgb]{0.459,0.562,0.841} $66.0\pm9.9$& \cellcolor[rgb]{0.62,0.38,0.66} $91.5\pm5.1$& \cellcolor[rgb]{0.569,0.569,0.569} $70.2\pm6.2$& $5.23\pm3.69$ & $2029\pm1157$ \\
L2 & \cellcolor[rgb]{0.3,0.66,0.42} $53.7\pm9.2$& \cellcolor[rgb]{0.32,0.45,0.8} $68.9\pm12.4$& \cellcolor[rgb]{0.676,0.471,0.71} $90.9\pm8.0$& \cellcolor[rgb]{0.52,0.52,0.52} $71.2\pm7.3$& $4.57\pm2.01$ & $1894\pm1014$ \\
L3 & \cellcolor[rgb]{0.509,0.762,0.593} $50.1\pm13.4$& \cellcolor[rgb]{0.511,0.605,0.856} $64.9\pm12.8$& \cellcolor[rgb]{0.954,0.926,0.959} $87.9\pm9.5$& \cellcolor[rgb]{0.69,0.69,0.69} $67.7\pm8.8$& $6.75\pm3.12$ & $2258\pm1102$ \\
L4 & \cellcolor[rgb]{0.718,0.863,0.767} $46.5\pm13.7$& \cellcolor[rgb]{0.861,0.888,0.959} $57.6\pm12.9$& \cellcolor[rgb]{0.685,0.486,0.718} $90.8\pm7.1$& \cellcolor[rgb]{0.821,0.821,0.821} $65.0\pm9.2$& $8.47\pm4.37$ & $2484\pm1217$ \\
L5 & \cellcolor[rgb]{0.916,0.959,0.93} $43.1\pm16.1$& \cellcolor[rgb]{0.918,0.934,0.976} $56.4\pm15.5$& \cellcolor[rgb]{0.954,0.926,0.959} $87.9\pm10.5$& \cellcolor[rgb]{0.942,0.942,0.942} $62.5\pm11.1$& $8.43\pm4.19$ & $2674\pm914$ \\
\midrule
\textit{All} & $\mathbf{49.3\pm13.5}$ & $\mathbf{62.8\pm13.5}$ & $\mathbf{89.8\pm8.3}$ & $\mathbf{67.3\pm9.1}$ & $\mathbf{6.69\pm3.87}$ & $\mathbf{2268\pm1106}$ \\
\bottomrule
\end{tabular*}%
}
\end{table*}

\begin{table*}[t]
\centering
\caption{Difficulty decay for \textbf{OpenClaw} agent runtime. Scores are percentages, reported as mean $\pm$ sample standard deviation across the 25 instances per level; \textit{All} pools the 125 instances. Each instance score averages three independent runs. Bold numbers mark pooled results. Resource statistics use successfully produced instances.}
\label{tab:decay-OpenClaw}
\setlength{\tabcolsep}{5pt}
\fittable{%
\begin{tabular*}{\textwidth}{@{\extracolsep{\fill}}lcccccc}
\toprule
\textbf{Level} & \textbf{Rule} & \textbf{Impl} & \textbf{Visual} & \textbf{Overall} & \textbf{Tokens (M)} & \textbf{Time (s)} \\
\midrule
L1 & \cellcolor[rgb]{0.3,0.66,0.42} $51.4\pm13.5$& \cellcolor[rgb]{0.32,0.45,0.8} $66.8\pm9.4$& \cellcolor[rgb]{0.62,0.38,0.66} $93.5\pm4.8$& \cellcolor[rgb]{0.52,0.52,0.52} $70.6\pm7.5$& $6.33\pm4.44$  & $2245\pm1378$ \\
L2 & \cellcolor[rgb]{0.516,0.765,0.599} $45.8\pm17.6$& \cellcolor[rgb]{0.546,0.632,0.866} $59.6\pm20.3$& \cellcolor[rgb]{0.767,0.621,0.792} $86.4\pm26.5$& \cellcolor[rgb]{0.686,0.686,0.686} $63.9\pm20.3$& $7.63\pm3.79$  & $2680\pm1324$ \\
L3 & \cellcolor[rgb]{0.766,0.886,0.806} $39.3\pm22.2$& \cellcolor[rgb]{0.793,0.833,0.939} $51.7\pm25.4$& \cellcolor[rgb]{0.954,0.926,0.959} $77.4\pm35.2$& \cellcolor[rgb]{0.876,0.876,0.876} $56.2\pm26.4$& $7.27\pm3.46$  & $2736\pm1254$ \\
L4 & \cellcolor[rgb]{0.808,0.907,0.841} $38.2\pm17.3$& \cellcolor[rgb]{0.871,0.896,0.962} $49.2\pm20.1$& \cellcolor[rgb]{0.838,0.736,0.855} $83.0\pm27.0$& \cellcolor[rgb]{0.861,0.861,0.861} $56.8\pm19.6$& $9.50\pm5.52$  & $3318\pm1867$ \\
L5 & \cellcolor[rgb]{0.916,0.959,0.93} $35.4\pm19.2$& \cellcolor[rgb]{0.918,0.934,0.976} $47.7\pm24.0$& \cellcolor[rgb]{0.954,0.926,0.959} $77.4\pm35.6$& \cellcolor[rgb]{0.942,0.942,0.942} $53.5\pm25.0$& $10.08\pm5.53$ & $3394\pm1598$ \\
\midrule
\textit{All} & $\mathbf{42.0\pm18.8}$ & $\mathbf{55.0\pm21.5}$ & $\mathbf{83.6\pm28.3}$ & $\mathbf{60.2\pm21.4}$ & $\mathbf{8.16\pm4.77}$ & $\mathbf{2875\pm1538}$ \\
\bottomrule
\end{tabular*}%
}
\end{table*}

\begin{table*}[t]
\centering
\caption{Difficulty decay for \textbf{OpenHands} agent runtime. Scores are percentages, reported as mean $\pm$ sample standard deviation across the 25 instances per level; \textit{All} pools the 125 instances. Each instance score averages three independent runs. Bold numbers mark pooled results. Resource statistics use successfully produced instances.}
\label{tab:decay-OpenHands}
\setlength{\tabcolsep}{5pt}
\fittable{%
\begin{tabular*}{\textwidth}{@{\extracolsep{\fill}}lcccccc}
\toprule
\textbf{Level} & \textbf{Rule} & \textbf{Impl} & \textbf{Visual} & \textbf{Overall} & \textbf{Tokens (M)} & \textbf{Time (s)} \\
\midrule
L1 & \cellcolor[rgb]{0.385,0.701,0.49} $45.4\pm16.7$& \cellcolor[rgb]{0.517,0.609,0.858} $57.0\pm17.0$& \cellcolor[rgb]{0.712,0.531,0.743} $84.3\pm14.9$& \cellcolor[rgb]{0.63,0.63,0.63} $62.3\pm14.8$& $9.30\pm6.22$  & $2173\pm1018$ \\
L2 & \cellcolor[rgb]{0.3,0.66,0.42} $46.9\pm13.5$& \cellcolor[rgb]{0.32,0.45,0.8} $62.5\pm15.3$& \cellcolor[rgb]{0.62,0.38,0.66} $87.5\pm13.5$& \cellcolor[rgb]{0.52,0.52,0.52} $65.6\pm12.4$& $11.70\pm8.02$ & $2704\pm1256$ \\
L3 & \cellcolor[rgb]{0.543,0.778,0.621} $42.6\pm25.3$& \cellcolor[rgb]{0.768,0.812,0.932} $50.0\pm27.3$& \cellcolor[rgb]{0.923,0.874,0.931} $77.0\pm27.6$& \cellcolor[rgb]{0.823,0.823,0.823} $56.5\pm25.1$& $14.29\pm7.24$ & $3065\pm1222$ \\
L4 & \cellcolor[rgb]{0.701,0.855,0.752} $39.8\pm19.9$& \cellcolor[rgb]{0.918,0.934,0.976} $45.8\pm22.3$& \cellcolor[rgb]{0.954,0.926,0.959} $75.9\pm20.2$& \cellcolor[rgb]{0.909,0.909,0.909} $53.9\pm19.3$& $16.03\pm8.27$ & $3433\pm1621$ \\
L5 & \cellcolor[rgb]{0.916,0.959,0.93} $36.0\pm22.2$& \cellcolor[rgb]{0.89,0.911,0.968} $46.6\pm25.2$& \cellcolor[rgb]{0.952,0.921,0.957} $76.0\pm22.4$& \cellcolor[rgb]{0.942,0.942,0.942} $52.9\pm21.8$& $18.11\pm6.64$ & $3717\pm1086$ \\
\midrule
\textit{All} & $\mathbf{42.2\pm20.0}$ & $\mathbf{52.4\pm22.5}$ & $\mathbf{80.1\pm20.6}$ & $\mathbf{58.2\pm19.5}$ & $\mathbf{13.88\pm7.85}$ & $\mathbf{3018\pm1353}$ \\
\bottomrule
\end{tabular*}%
}
\end{table*}

\section{Game Replication Prompt}
\label{app:example}

\begin{promptbox}{Game Replication System Prompt}
\scriptsize
\textbf{\S0. Task}

Your task is to black-box replicate a web game.

\texttt{game-server} is a standalone HTTP service that loads the original game in an internal headless browser and exposes it to you only through the screenshot, click, keyboard, scroll, drag, hover, and reset interfaces. It is your \emph{only} channel for observing the original game: you cannot read its source code, DOM, or any files. You can rely only on the screenshot pixels it returns and the visible results after your actions to understand the original game's UI, interactions, rules, and termination conditions.

Based on this, you must implement a runnable, interactive HTML replica page. \texttt{game-server} can likewise load and screenshot your replica for self-check comparison (interfaces in \S1).

Instance information is appended at the end, containing at least three values, referred to throughout by placeholders:

\begin{itemize}
  \item \texttt{GAME\_URL}: the HTTP address of \texttt{game-server}.
  \item \texttt{OUTPUT\_HTML}: the path of the replica HTML file you must generate (must be a \texttt{.html} file, not a directory).
  \item \texttt{REPLICA\_PATH}: the relative path \texttt{game-server} uses when rendering your replica; use the value from the appended information directly, do not construct it yourself.
\end{itemize}

\textbf{You must build the replica page \texttt{OUTPUT\_HTML}}, and also generate the document \texttt{docs/game\_spec.md} (see \S4). In addition, you must save screenshots to \texttt{screenshots/original/} (exploring the original game) and \texttt{screenshots/replica/} (self-checking the replica), with filenames reflecting order, using the \texttt{cycle\_<cycle>\_step\_<step>.png} format (\texttt{cycle} = which ``explore $\to$ implement $\to$ self-check'' loop; \texttt{step} = which action within that loop; both zero-padded from 001, e.g., \texttt{cycle\_001\_step\_003.png} = loop 1, step 3). No process files other than the above may be produced.

Output structure (relative to the task working directory; paths must be exact --- if \texttt{docs/game\_spec.md} is misplaced or renamed, scoring silently skips it and gives a direct zero):

{\scriptsize
\begin{verbatim}
.
|-- docs/
|   `-- game_spec.md
|       # scored document, see S4; path must not change
|-- screenshots/
|   |-- original/         # game screenshots
|   `-- replica/          # self-check screenshots
`-- <OUTPUT_HTML>         # deliverable HTML file
\end{verbatim}
}

\textbf{\S1. Available Interfaces}

You may call only the following \texttt{game-server} interfaces; do not call any other.

\begin{itemize}
  \item Click body: \texttt{\{"x":100,"y":200\}} (pixel coordinates estimated from the screenshot). Left button by default; for right button add \texttt{"button":"right"} (for right-click menus / marking interactions).
  \item Keyboard body: \texttt{\{"key":"ArrowLeft"\}} (e.g., ArrowUp/Down/Left/Right/Enter/Space) or \texttt{\{"text":"abc"\}}.
  \item Scroll body: \texttt{\{"deltaY":300\}} scrolls down 300px (negative for up; \texttt{deltaX} for horizontal). Add \texttt{"x"}/\texttt{"y"} to move the cursor first, for scrolling a specific region.
  \item Drag body: \texttt{\{"x1":100,"y1":200,"x2":300,"y2":200\}} (press at start, move to end, release; also accepts \texttt{\{"from":...,"to":...\}}; add \texttt{"button":"right"} for right-drag). For slider / line-drawing / piece-dragging.
  \item Hover body: \texttt{\{"x":100,"y":200\}} (move cursor only, no click; for observing hover appearance).
  \item All screenshot responses are PNG.
\end{itemize}

{\scriptsize
\begin{verbatim}
Original game  (base = $GAME_URL):
  GET  /screenshot/original
  POST /action  /keyboard
       (return the post-action screenshot)
  POST /scroll  /drag  /hover  /reset

Replica page (base = $GAME_URL;
              add ?path=$REPLICA_PATH to each):
  GET  /screenshot/replica
  POST /action/replica   /keyboard/replica
  POST /scroll/replica   /drag/replica
  POST /hover/replica    /reset/replica
\end{verbatim}
}

Examples:

{\scriptsize
\begin{verbatim}
curl -s "$GAME_URL/screenshot/original" -o o.png
curl -s -X POST "$GAME_URL/action" \
  -d '{"x":100,"y":200}'
curl -s -X POST "$GAME_URL/keyboard" \
  -d '{"key":"ArrowLeft"}'
curl -s -X POST "$GAME_URL/scroll" \
  -d '{"deltaY":300}'
curl -s -X POST "$GAME_URL/drag" \
  -d '{"x1":100,"y1":200,"x2":300,"y2":200}'
curl -s -X POST "$GAME_URL/hover" \
  -d '{"x":100,"y":200}'
\end{verbatim}
}

Original-game interfaces change the original page state until reset. Replica interfaces do not modify your files; they only make \texttt{game-server} load the corresponding HTML and screenshot it. If keyboard has no effect, click the screen to gain focus and retry.

\textbf{Operation tips}: prefer the interface closest to the original interaction --- use \texttt{/drag} for draggable things, \texttt{/scroll} for scrollable ones, \texttt{/hover} to inspect hover states; ``select then move'' can also be expressed with two clicks (click the object, then the target). If a screenshot looks cropped or seems to contain more content, try \texttt{/scroll} first; if the view does not change after scrolling, it is a single screen --- troubleshoot per \S7 (reset / re-screenshot), do not keep scrolling in vain. If an action requiring a missing interface truly cannot be completed, judge whether it is blocked per \S7.

\textbf{\S2. Hard Boundaries (Black-Box)}

\textbf{Your only source of information is screenshots and interaction results.} The original game's source code, tests, specification, rubric, and answers are not in your container; you can only screenshot and operate it through \texttt{game-server} (\S1). Your working directory holds only this task's outputs --- do not rummage elsewhere for the original or the answers. Even if something is readable somewhere, you must not use it --- all rules must come from your own observation.

\textbf{In particular: do not presume gameplay from the game's name, file names, or common sense.} Even if the screen looks like a game you know, its layout, rules, and win/lose conditions must still be observed and verified one by one; they cannot be asserted from prior knowledge.

\textbf{Do not hard-code any win/lose condition, hidden answer, scoring rule, or key gameplay logic that you have not observed and confirmed.}

\textbf{\S3. Replica Implementation Contract (must follow)}

Evaluation reads your replica's code and DOM to verify state, so the game state must be carried in a machine-readable way on the DOM and a global state object, not only in pixels. Both of the following must hold.

\textbf{\S3.1 Per-element DOM; do not draw everything on a canvas / bare SVG.}
\begin{itemize}
  \item Every state-bearing unit in the game area (cell, piece, tile, slider, line, marker, etc.) must be a real DOM element (\texttt{<div>}/\texttt{<button>}/\texttt{<span>}, etc.), whose current state is reflected through its own tag / \texttt{class} / \texttt{data-*} (color, orientation, selected, completed, coordinates).
  \item Do not draw the whole board in a single \texttt{<canvas>}, or render it as a single \texttt{<svg>} without locatable child elements. Purely decorative details that do not participate in adjudication are unrestricted.
  \item Status text, buttons, and scores likewise use ordinary DOM elements; bind click targets on the corresponding DOM elements where possible.
\end{itemize}

\textbf{\S3.2 Expose \texttt{window.\_\_GAME\_STATE\_\_}.} It must exist right after page load, \textbf{update synchronously after every action}, and fully reflect all currently visible objects and key dynamic state (board dimensions, each object's coordinates and attributes, counters, win/end flags, etc.), with \textbf{0-based} coordinates (top-left \texttt{(0,0)}). Field names must be intuitive and stable, and must stay consistent with the display / DOM --- state must not disagree with the screen. This applies only to your own replica and does not violate \S2.

Example (fields vary by game): \texttt{\{rows, cols, player:\{row,col\}, boxes:[\{row,col\}], goals:[\{row,col\}], walls:[[row,col]], moves, won\}}

\textbf{\S4. \texttt{docs/game\_spec.md} (the scored document)}

This file is compared line by line against the ground-truth rubric to compute the rule score, so it is \textbf{not a game introduction but a verifiable checklist}. It must align with the evaluation and use a fixed taxonomy of \textbf{five mandatory categories plus one optional category}, each forming its own section with continuous per-section numbering:

\begin{itemize}\setlength{\itemsep}{2pt}
  \item \texttt{Visual} --- static appearance (background, layout, status bar, buttons, board position/size; the look of each object type; selected/hover/disabled looks) plus any transitions (move/fall/flip/flash/fade) and their timing/timers; if there is none, write ``instant switch, no transition.''
  \item \texttt{Legal action} --- input method; the visible state change after each legal action; forward effects of Undo/Restart/Next.
  \item \texttt{State} --- how moves/score/lives/remaining change; board invariants and state value domains.
  \item \texttt{Boundary} --- illegal/invalid actions: no state change, no counting, and no penalty (wall, blank space, locked cell, acting after game end).
  \item \texttt{Terminal lock} --- win/lose/draw conditions and text; whether the board locks after game end; whether a restart entry appears.
  \item \texttt{Level layout} --- (optional) coordinate system and key positions for fixed levels/puzzles; omit the whole section if not applicable.
\end{itemize}

The five mandatory categories (Visual, Legal action, State, Boundary, Terminal lock) must always form a section regardless of game type (if a category has no content, keep the header and write one explanatory line, e.g., ``no failure condition observed''). Level layout appears only when such an element exists. \textbf{Only these six categories may be used; do not add other categories (e.g., sound effects).}

\textbf{Every rule must carry a confidence tag in square brackets after its number} (because your rules come from observation and are not guaranteed to be ground truth):
\begin{itemize}
  \item \texttt{[confirmed]}: stably verified through observation or active experiment, with screenshot evidence. Must be implemented.
  \item \texttt{[inferred]}: reasonably supported but not fully verified. May be implemented as a plausible assumption to keep the page playable, but must not be disguised as confirmed.
  \item \texttt{[unknown]}: not observed or not verifiable. Do not hard-code its win/lose outcome, answer, scoring, or critical logic.
\end{itemize}

Authoring rules:
\begin{enumerate}
  \item \textbf{Granularity (most important): one element / one state / one behavior per entry; do not split to a single attribute.} The atomic unit is ``one independently verifiable object, state, or behavior,'' not ``one CSS property.''
  \begin{itemize}
    \item \emph{Do split} (each part can be judged true/false independently): different elements get separate entries; different interaction states of the same element (default/hover/selected/disabled) get separate entries; the positive effect and the no-op case of the same action get separate entries (positive $\to$ Legal action/State, no-op $\to$ Boundary).
    \item \emph{Do not split} (a group of static style sub-attributes of the same element in the same state, always appearing and judged together): put them in one entry and list sub-attributes with semicolons. \textbf{One attribute per entry is forbidden} (e.g., a button's background/text color, font weight, and corner radius form one entry, not four).
    \item Criterion: if two points ``are right together or wrong together in the same element and the same screenshot,'' merge them; split only when the element/state/action changes and each can be judged independently.
  \end{itemize}
  \item \textbf{Stable numbering}: number continuously from 1 within each category, prefixed by the category name (e.g., Visual-1, Visual-2\dots State-1\dots); do not change a number once assigned.
  \item \textbf{Verifiable, not vague}: descriptions like ``nice-looking UI'' or ``moderate difficulty'' are forbidden; write observable, verifiable facts (e.g., ``clicking the round button at the bottom-right increments Moves by 1'').
  \item \textbf{No implementation details}: write from user-observable visual/interactive behavior; do not reference class names, ids, \texttt{data-*}, function names, variable names, \texttt{window.\_\_GAME\_STATE\_\_}, or any non-visual source.
  \item \textbf{Approximate values allowed}: write colors as approximations with a note (e.g., ``background approximately \texttt{\#e7ede4}''); give concrete values or ranges for sizes/durations.
  \item \textbf{Split positive and negative}: the positive effect of an action goes to Legal action/State, the no-op case to Boundary, under different numbers. Negative rules are key to catching missing guards --- do not write only the positive.
  \item \textbf{Hidden mechanics} get their own entry: state whether they are visible, how they are triggered, the effect after triggering, and whether they carry a counter/state penalty; file them under the appropriate category; if unconfirmed, tag inferred/unknown.
  \item \textbf{Non-redundant --- no derived results}: write only raw rules/facts, not ``emergent results'' derivable from other entries. Typical: do not write a complete solution (step-by-step moves) --- it follows from the movement/blocking/layout rules; solvability may be expressed by at most one invariant entry.
  \item \textbf{Deduplicate}: each fact is written once in its most appropriate category, not repeated across categories (e.g., board dimensions go only in Level layout, not repeated in Visual and State).
\end{enumerate}

Excerpt example (Sokoban, to show format / granularity / categorization / confidence, not asking you to replicate Sokoban):

{\scriptsize
\begin{verbatim}
## Visual
- Visual-1 [confirmed] Page background light beige,
  approx #e7ede4.
- Visual-2 [confirmed] Board is 6 cols x 5 rows,
  with square cells, centered.
- Visual-3 [confirmed] Wall cell: dark-gray solid
  square; no inner mark.
- Visual-4 [confirmed] Goal cell: light base with a
  central red dot.
- Visual-5 [confirmed] Player: green circle.
- Visual-6 [confirmed] Box: tan square with a dark
  outline.
- Visual-7 [inferred] Box border turns red on a goal
  (seen once).
## Legal action
- Legal action-1 [confirmed] Arrow keys move the player
  one cell (if empty).
- Legal action-2 [confirmed] Player pushes an adjacent
  box when the cell beyond is empty; both move one
  cell.
## State
- State-1 [confirmed] Each move increments "Moves";
  Restart zeroes it.
## Boundary
- Boundary-1 [confirmed] Moving into a wall, or a box
  backed by a wall/box, does not move or change
  "Moves".
## Terminal lock
- Terminal lock-1 [confirmed] All boxes on goals ->
  win;
  shows "You Win!".
- Terminal lock-2 [inferred] After winning, the board
  locks and keys are inert.
## Level layout
- Level layout-1 [confirmed] Player (1,1); boxes at
  (2,2) and (3,2); goals at (4,3) and (4,4).
  A wall ring surrounds the board.
\end{verbatim}
}

\textbf{\S5. Execution Workflow}

Proceed as explore $\to$ implement $\to$ self-check; when you find an inconsistency, go back to explore more and edit the same \texttt{OUTPUT\_HTML}, without maintaining multiple versions.

\textbf{Explore.} First \texttt{/reset} + screenshot to confirm the initial screen is legible (blank / failed load: troubleshoot per \S7). Dismiss any overlay / popup / start screen to reach the main view. Then actively design multi-step operations to observe: initial layout and appearance; how the target and linked elements change after one action; whether legal / illegal / boundary actions each take effect; whether reset returns to the initial state; whether you can reach and observe the end state and restart behavior; if randomness is suspected, reset and re-verify several times. \textbf{Actively perform illegal boundary actions} (click blank space, repeat a click, wrong timing, click a completed area) and \textbf{controlled experiments} (change only one variable and compare screenshots). Save each key screenshot to \texttt{screenshots/original/} with ordered names (e.g., \texttt{cycle\_001\_step\_003.png}). Stop when screenshots start repeating and yield no new information. If all exploration screenshots are identical, the actions may have had no effect --- troubleshoot and re-explore before implementing.

\textbf{Induce.} Write observations into \texttt{docs/game\_spec.md} (per the \S4 format: the six categories, appropriate granularity, stable numbering, confidence tags).

\textbf{Implement.} Create the parent directory of \texttt{OUTPUT\_HTML} and generate the replica. Prefer a self-contained static page (HTML/CSS/vanilla JS), implemented with per-element DOM per \S3. Implement all selected confirmed items; implement inferred items only when needed for playability, tagged as guesses; do not hard-code unobserved critical logic. The UI must match the observed layout, size, palette, shapes, text, typographic feel, and the end / restart screen.

\textbf{Self-check.} You must render the replica through \texttt{game-server} (do not only open it locally). Screenshot the replica's initial and post-action states (save to \texttt{screenshots/replica/}, also ordered) and compare them item by item (by \texttt{game\_spec} numbering) against the corresponding original screenshots --- Visual for appearance and animation, Legal action / State for action results and counter linkage, Boundary for whether illegal actions are equally inert, Terminal lock for end state and reset, Level layout for coordinates. For non-deterministic elements, compare only behavioral categories. ``Close'' does not count as ``consistent.'' Before finishing, also \texttt{Read} your own \texttt{OUTPUT\_HTML} once to statically confirm it satisfies \S3.

\textbf{\S6. Convergence and Stopping}
\begin{itemize}
  \item \textbf{Budget}: $\leq$50 screenshots each for the original game and the replica; $\leq$3 ``edit page then re-check'' loops. Do not loop indefinitely.
  \item \textbf{Continue} (when new information or a concrete fix is still possible): confirmed items inconsistent (especially unplayable page / key interaction failing), observed UI not yet replicated, key gameplay still unknown but explorable, inferred items pending and testable.
  \item \textbf{Stop}: all implemented items have consistent self-check evidence, or \S7 blocked is met.
  \item Before stopping, honestly list remaining inferred / unknown; do not fabricate a consistent conclusion.
\end{itemize}

\textbf{\S7. Blocked and Troubleshooting}
\begin{itemize}
  \item Do not ask the user questions during execution.
  \item Common fixes:
  \begin{itemize}
    \item Blank screenshot $\to$ wait / reset / re-screenshot.
    \item Click has no effect $\to$ change coordinates, click the element center, check for an illegal state.
    \item Replica won't open $\to$ check \texttt{OUTPUT\_HTML} exists and is a file, and \texttt{REPLICA\_PATH} is correct.
    \item Keyboard has no effect $\to$ click the screen to gain focus, then send keys.
  \end{itemize}
  \item Reporting blocked requires all of: the same problem prevents progress, at least three essentially different recovery strategies tried, each attempt evidenced.
\end{itemize}

\textbf{\S8. Final Reply}

Report briefly, one of two:
\begin{itemize}
  \item \textbf{Done}: state that replication is complete, giving \texttt{OUTPUT\_HTML}, \texttt{REPLICA\_PATH}, and a short self-check summary.
  \item \textbf{Blocked}: summarize the blocking cause, the recovery strategies tried, and the evidence.
\end{itemize}
\end{promptbox}

\end{document}